\documentclass[10 pt,journal]{IEEEtran}
\IEEEoverridecommandlockouts

\usepackage{times}
\usepackage{import}
\usepackage{todonotes}
\usepackage{amsmath}
\usepackage{mathtools}
\usepackage{amsmath,bm}
\usepackage{relsize}
\usepackage{hyperref}
\usepackage{graphics}
\usepackage{subcaption}
\usepackage{amssymb}
\usepackage{mathrsfs}
\usepackage{paralist}
\usepackage{comment}
\usepackage{rotating}
\usepackage{multirow}
\usepackage{colortbl}
\usepackage{tabularx}
\usepackage{makecell}
\usepackage{soul}
\usepackage{caption}
\usepackage{etoolbox}

\definecolor{Gray}{gray}{0.95}

\let\mybibitem\bibitem
\renewcommand{\bibitem}[1]{%
  \ifstrequal{#1}{nature}
    {\color{blue}\mybibitem{#1}}
    {\color{black}\mybibitem{#1}}%
}

\title{ \LARGE \bf Teleoperation of Humanoid Robots: A Survey}

\author
{Kourosh Darvish$^{1}$*, Luigi Penco$^{2}$, Joao Ramos$^{3}$,  Rafael Cisneros$^{4}$,\\ Jerry Pratt$^{2}$, Eiichi Yoshida$^{6}$, Serena Ivaldi$^{5}$, Daniele Pucci$^{1}$
\thanks{\hspace{-0.25cm}$^{1}$ Artificial and Mechanical Intelligence, Center for Robotics and Intelligent Systems, Istituto Italiano di Tecnologia, Genoa, Italy,
        {\tt\small Kourosh.Darvish@gmail.com,~Daniele.Pucci@iit.it
        }\newline
        $^{2}$  Florida Institute for Human and Machine Cognition, 40 S Alcaniz St, Pensacola, FL 32502, United States
        {\tt\small lpenco@ihmc.org,
        jpratt@ihmc.org
        }\newline
        $^{3}$ Department of Mechanical Science and Engineering, University of Illinois at Urbana-Champaign, USA,
        {\tt\small jlramos@illinois.edu
        }\newline
        $^{4}$  CNRS-AIST JRL (Joint Robotics Laboratory), IRL; National Institute of Advanced Industrial Science and Technology (AIST), Department of Information Technology and Human Factors, Tsukuba, Japan,
        {\tt\small rafael.cisneros@aist.go.jp
        }\newline
        $^{5}$ Inria, Loria, Universit\'e de Lorraine, CNRS, Nancy, France,
        {\tt\small serena.ivaldi@inria.fr
        }\newline
        $^{6}$ Department of Applied Electronics, Faculty of Advanced Engineering, Tokyo University of Science, Japan,
        {\tt\small eiichi.yoshida@rs.tus.ac.jp
        }\newline
        $^{*}$ Corresponding author 
        }
\thanks{This work has received funding from the European Union's Horizon 2020 research and innovation programmes under grant agreement No. 731540 (An.Dy), No. 869855 (SoftManBot), No. 101070596 (EuRobin), \& the Italian National Institute for Insurance against Accidents (INAIL) ergoCub project.
}
}

\date{}

\begin{document} 

\bstctlcite{IEEEexample:BSTcontrol}



\maketitle


\thispagestyle{plain}
\pagestyle{plain}


\begin{abstract}
Teleoperation of humanoid robots enables the integration of the cognitive skills and domain expertise of humans with the physical capabilities of humanoid robots.
The operational versatility of humanoid robots makes them the ideal platform for a wide range of applications  when teleoperating in a remote environment. 
However, the complexity of humanoid robots imposes challenges for teleoperation, particularly in unstructured dynamic environments with limited communication.
Many advancements have been achieved in the last decades in this area, but a comprehensive overview is still missing.
This survey paper gives an extensive overview of humanoid robot teleoperation, presenting the general architecture of a teleoperation system and analyzing the different components.
We also discuss different aspects of the topic, including technological and methodological advances, as well as potential applications.
A web-based version of the paper can be found at \url{https://humanoid-teleoperation.github.io/}.
  
\end{abstract}
 
\begin{IEEEkeywords}
Humanoid robot; Teleoperation.
\end{IEEEkeywords}
 

\section{Introduction}
\label{sec:Introduction}

\begin{figure}[!t]
 \setlength\belowcaptionskip{-0.7\baselineskip}
    \centering
    \includegraphics[width=0.8\columnwidth]{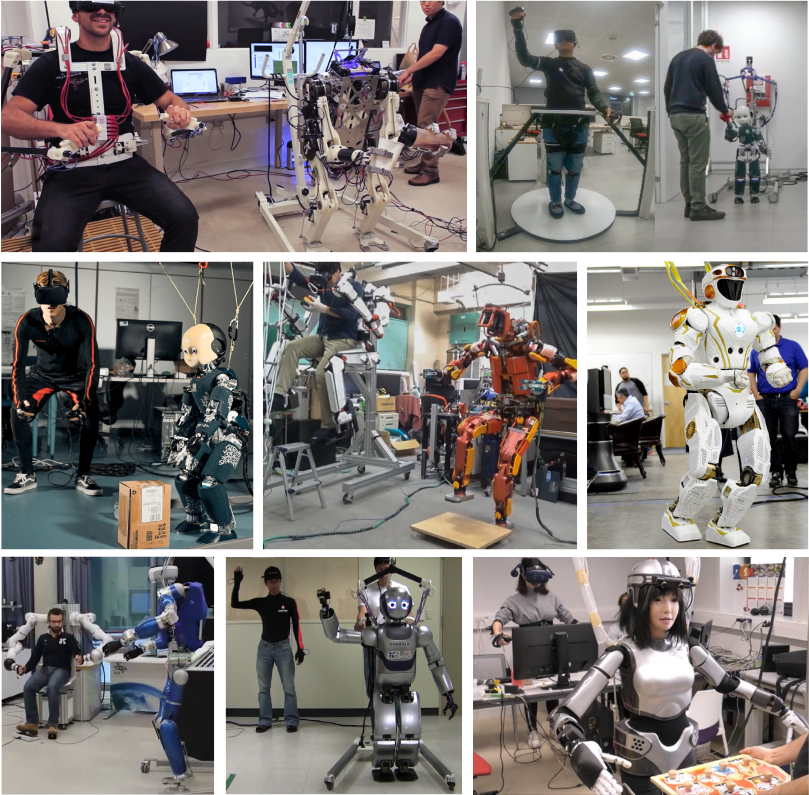}
    \caption{ Examples of humanoid robot teleoperation; from top left to bottom right corner: \cite{ramos2018}, \cite{darvish2019}, \cite{penco2019}, \cite{ishiguro2020bilateral}, \cite{jorgensen2019}, \cite{abi2018}, \cite{kim2013}, \cite{Cisneros2022Team}.}
    \label{fig:intropic}
\end{figure}

There are many situations and environments where we need robots to replace humans at the site. 
Despite the recent progress in robot cognition based on AI techniques, fully autonomous solutions are still far from producing socially and physically competent robot behaviors; that is why teleoperating robots (Fig.~\ref{fig:intropic}) acting as physical avatars of human workers at the site is the most reasonable solution.
In environments like construction sites, chemical plants, contaminated areas and space, teleoperated robots could be extremely valuable, relieving humans from any potential hazard.
Contrary to other conventional robotic platforms, humanoids' structure is a better fit for environments and tasks that are designed for and performed by humans. The operational versatility of these robots makes them suitable for work activities that require a variety of complex mobility and manipulation skills, such as inspection, maintenance, and interaction with human operators.
In certain contexts such as telenursing, where human subjects are expected to interact with a teleoperated robot, the human-likeness factor is important since it increases the acceptability, social closeness, and legibility of its intentions~\cite{dragan2013a}.

In the literature, different attempts have been made to \textit{``deploy the senses, actions, and presence of a human to a remote location in real-time, leading to a more connected world''}~\cite{AnaAvatarXprize}.
Inspired by a visit to the Tachi Lab, the XPRIZE Foundation has recently launched the ANA Avatar XPRIZE global competition~\cite{AnaAvatarXprize}.
Previously, in response to the 2011 Fukushima Daiichi nuclear disaster, the DARPA Robotics Challenge (DRC) was launched to promote innovation in human-supervised robotic technology.
In space applications, rovers and mobile manipulators were teleoperated from aboard the International Space Station (ISS), in the context of METERON and Kontur-2 projects~\cite{kontur-meteron}. In 2019, the humanoid Skybot F-850 was rocketed to the ISS~\cite{skybot2019}; however, it turned out to have a design that did not work well, demonstrating that there is still work to do to get humanoids into space.

Humanoid robot teleoperation involves many multidisciplinary and interleaved challenges, ranging from dynamics and control to communication and human psychophysiology. Uniquely, due to their resemblance to human appearance, societal expectations are high as well;
they are expected to do a wide range of tasks that are not expected from other types of robots. They are highly redundant with nonlinear, hybrid, and underactuated models.
While doing dynamic and agile motions with the feet like walking, running, or stepping over obstacles, they are supposed to perform dexterous power and precision manipulation.
At the same time,
they are expected to work alongside humans, be safe, friendly, and socially interact with others.
On the other hand, teleoperation interfaces and techniques should be designed such that the human operator receives minimal, effective, and informative haptic feedback from the humanoid robot, to cover for human errors, overcome communication delays, and above all, be telepresent. Along with these challenges, the field is new and due to its high resource demand for development, not many laboratories have been working on it.

Many efforts from the robotics community have been devoted to studying humanoid robots, teleoperation, evaluation metrics, or human-robot interaction.
Among them, the book on humanoid robotics \cite{goswami2019humanoid} studied comprehensively different aspects of humanoid robots, including their history, design, mechanics, control, simulation, and interaction. 
Several survey papers likewise studied specific aspects of humanoid robots, for example humanoid dynamics~\cite{sugihara2020survey}, control \cite{yamamoto2020survey}, motion generation \cite{Tazaki2020survey}, or robot teleoperation interface design and metrics \cite{de2009survey}.
A primary work on bilateral teleoperation techniques has been presented by \cite{hokayem2006bilateral} as well. 
Another seminal survey \cite{Goodrich2007Survey} covers many aspects of interactive robots, including their design, autonomy level, and human factors that helped us in articulating the current manuscript.
Another interesting survey \cite{goodrich2013teleoperation} highlights several aspects of humanoid teleoperation and autonomy.
However, \cite{goodrich2013teleoperation} is a decade old, and an up-to-date survey on the topic is missing, especially considering that humanoid teleoperation is a far-from-solved challenge and a highly active field of research where new solutions are proposed each year.
Following the workshop in~\cite{workshop2019}, this survey paper presents the latest results in humanoid robot teleoperation and draws in detail the challenges that the research community faces to effectively deploy such systems toward real-world scenarios.

Starting from what emerged from the workshop, we conducted a survey on teleoperation of humanoid robots. We present here the systems and devices that have been adopted so far to teleoperate humanoids (Sec.~\ref{sec:TeleoperationSystemDevices}) and how these robots have been modeled, retargeted, and controlled (Sec.~\ref{sec:RobotControl}). We also examine a
promising case of teleoperation in which the robot assists the user in accomplishing a desired task (Sec.~\ref{sec:AssistedTele}). 
Later, we discuss complications along with some compensating solutions that arise due to non-ideal communication channels (Sec.~\ref{sec:CommunicationChannel}).
We explain the evaluation of teleoperation systems prior to development to meet the users' needs (Sec.~\ref{sec:EvaluationMetrics}). 
Finally, discussions on current and potential applications and the associated challenges follow (Sec.~\ref{sec:Applications}).  

\section{Teleoperation System and Devices}
\label{sec:TeleoperationSystemDevices} 

\begin{figure}[!t]
 \setlength\belowcaptionskip{-0.7\baselineskip}

\centering
\includegraphics[width=\columnwidth]{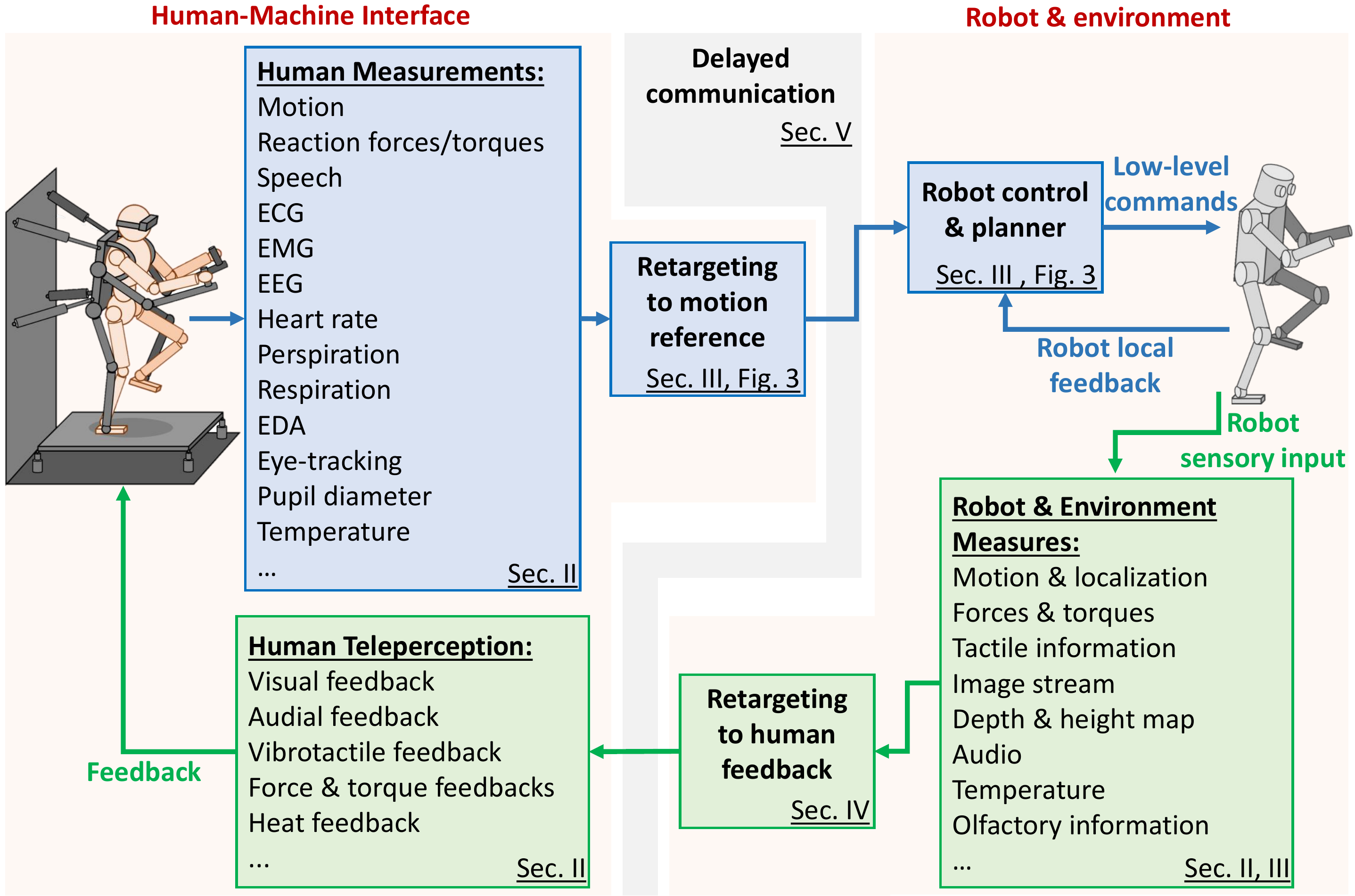}
\caption{Schematic architecture for teleoperating a humanoid.}
\label{fig:teleoperation_architecture}
\end{figure}

In the literature, the terms teleoperation and telexistence have been used indistinctly in different contexts. Telexistence refers to the technology that allows human to virtually exist in a remote location through an avatar, experiencing real-time sensations from the remote site \cite{tachitelexistence}. Both the remote environment and the avatar can be real or virtual, but in this article we only consider a real environment and a surrogate humanoid robot as avatar. Telexistence has also been referred to as telepresence in the literature \cite{tachitelexistence}.
The concept of teleoperation, on the other hand, still refers to a human operator remotely controlling a robot, but the focus is mainly put on performing  tasks that require high dexterity in the remote location.
We use these terms interchangeably throughout this survey.
From another perspective, the teleoperation setup represents an interactive system where the robot imitates the human's actions to reach a common objective \cite{Goodrich2007Survey}.

In a teleoperation setup, the user is the person who teleoperates the humanoid robot and identifies the teleoperation goal, i.e. intended outcome, through interfaces. 
The interfaces are the means of the interaction between the user and the robot.
The nature of the exchanging information, constraints, the task requirements, and the degree of shared autonomy determine different preferences on the interface modalities, according to specific metrics which will be discussed later.
Moreover, the choice of the interfaces should make the user feel comfortable, hence enabling a natural and intuitive teleoperation.

\subsection{Teleoperation Architecture}
\label{sec:TeleoperationArchitecture }
Fig.~\ref{fig:teleoperation_architecture} shows a schematic view of the architecture for teleoperating a humanoid robot. 
First, human kinematic and dynamic information are measured and transmitted to the humanoid robot motion for teleoperation.
More complex retargeting methods (i.e., mapping of the human motion to the robot motion), employed in assisted teleoperation systems (Sec.~\ref{sec:AssistedTele}), may need the estimation of the user reaction forces/torques.
There are cases in which also physiological signals are measured in order to estimate the psychophysiological state of the user, which can help enhancing the performance of the teleoperation.
On the basis of the estimated states, the retargeting policy is selected and the references are provided to the robot accordingly.
Teleoperation systems are employed not only for telemanipulation scenarios but also for social teleinteractions (i.e. remotely interacting with other people). In this case, the robot's anthropomorphic motion and social cues such as facial expressions can enhance the interaction experience. Therefore, rich human sensory information is indispensable.

To effectively teleoperate the robot, the user should make proper decisions; therefore he/she should receive various feedback from the remote environment and the robot.
In many cases, the sensors for perceiving the human data and the technology to provide feedback to the user are integrated together in an interface.
In the rest of the section, we will discuss  different available interfaces and sensor technologies in teleoperation scenarios,
whereas their design and evaluation will be discussed later in Sec.~\ref{sec:EvaluationMetrics}.

The retargeting block (Fig.~\ref{fig:teleoperation_architecture}) maps human sensory information to the reference behavior for the robot teleoperation, hence the human can be considered as master or leader, and the robot as slave or follower.
We can discern two retargeting strategies: unilateral and bilateral teleoperation. In the unilateral approach, the coupling between the human and robot takes place only in one direction.
The human operator can still receive haptic feedback either as a kinesthetic cue not directly related to the contact force being generated by the robot or as an indirect force in a passive part of her/his body not commanding the robot.
But in bilateral systems, human and humanoid robot are directly coupled. The choice of bilateral retargeting depends on the task, communication rate, and degree of shared autonomy, as will be discussed~in~Sec.~\ref{sec:AssistedTele}.

The human retargeted information, together with the feedback from the robot, are streamed over a communication channel that could be non-ideal. In fact, long distances between the operator and the robot or poor network conditions may induce delays in the flow of information, packet loss and limited bandwidth, adversely affecting the teleoperation experience.
Sec.~\ref{sec:CommunicationChannel} details the approaches in the literature to teleoperate robots in such conditions.
Finally, the robot local control loop
generates
the low level commands- i.e., joints position, velocity, or torque references- to the robot, taking into account the human references (Sec.~\ref{sec:RobotControl}).

\subsection{Human Sensory Measurement Devices}
\label{sec:HumanSensoryMeasurementsDevices}
In this section, we provide a detailed description of the technologies available to sense various human states,
including the advances to measure the human motion, the physiological states, and the interaction forces with the environment.
We report in TABLE \ref{tab:teleoperation} the various measurement devices adopted so far for humanoid robot teleoperation.

\subsubsection{Human kinematics and dynamics measurements}
To provide the references for the robot motion, we need to sense human intentions.
For simple teleoperation cases, the measurements may be granted through simple interfaces such as keyboard, mouse, or joystick commands \cite{ando2020master}.
However, for a more complex system like a humanoid robot, those simple interfaces may not be enough, especially when the user wants to exert a high level of control authority over the robot. Therefore, the need for natural interfaces for effectively commanding the robot arises. 
We can consider solutions benefiting from the similarity of the human and the humanoid robot anthropomorphic geometries, i.e., providing the references to the robot limbs with spatial analogies to those of the human (natural mapping).
Therefore, the need to measure the human kinematics emerges.

Different technologies have been employed in the literature to measure the motion of the main human limbs, such as legs, torso, arms, head, and hands. 
An ubiquitous option are the Inertial Measurement Unit (IMU)-based wearable technologies. In this context,
two cases are possible,
a segregated set of IMU sensors are used throughout the body to measure the human motion 
or an integrated network of sensors is used throughout the human body \cite{darvish2019, penco2019}.
The former case normally provides the raw IMU values, while the latter provides the information of the human limb movements.
In the second case,
a calibration process computes the sensors transformations with respect to body segments \cite{roetenberg2009xsens}. This technology is especially of interest because of the high accuracy and frequency of the retrieved human motion information
without the occlusion problem.
However, its accuracy may suffer from disturbances caused by the magnetic field and displacement of the sensors with respect to their initial emplacement.
Yet, another recent wearable technology to capture the hand pose is a stretchable soft glove embedded with distributed capacitive sensors \cite{glauser2019interactive}.
A review about the textile-based wearable technology used to estimate the human kinematics is provided in \cite{wang2020textile}.

Optical sensors are another technology used to capture the human motion, with active and passive variants. In the active case, the reflection of the pattern is sensed by the optical sensors. Some of the employed technologies include depth sensors and optical motion capture systems.
In the case of passive sensors, RGB monocular cameras (regular cameras) or binocular cameras (stereo cameras) are used to track the human motions. Thanks to the optical sensors, a skeleton of the human body is generated and tracked in 2-D or 3-D Cartesian space. The main problems with these methods are the occlusion and the low portability of the setup.

To track the users' motion in bilateral teleoperation scenarios, exoskeletons are often used.
In this case, the exoskeleton model and the encoder data are fed to the forward kinematics to estimate the human link's poses and velocities~\cite{ramos2019dynamic, ishiguro2020bilateral}.

The previously introduced technologies can be used to measure the human gait information with locomotion analysis.
Conventional or omnidirectional treadmills are employed in the literature for this purpose~\cite{ elobaid2018a}. While the treadmill can be used for even terrains, it would not work to retarget locomotion on uneven terrains.
To respond to this shortcoming, a cockpit-like teleoperation setup has been recently proposed~\cite{ishiguro2020bilateral}.
To estimate the interaction forces between the human and the teleoperation setup, force-torque sensors measuring the human wrenches can be integrated in ground plates, shoes, or exoskeletons.
Richer information can be obtained by distributed capacitance sensors that measure the pressure manifold. 


\begin{table*}
  \setlength\belowcaptionskip{-0.7\baselineskip}

\setlength\tabcolsep{1.5pt} 
\centering
\arrayrulecolor{black}
\caption{Main works and technologies related to the teleoperation of humanoids.}
\begin{tabular}{|c|c|c|c|c|c|c|c|c|c|c|c|c|c|c|c|c|c|c|c|c|c|c|c|c|c|} 
\hline
\multirow{4}{*}{ ref} &
\multirow{4}{*}{ robot} &
\multicolumn{11}{c|}{teleoperation devices} &
\multicolumn{4}{c|}{retargeting \& planning} &
\multicolumn{3}{c|}{stabilizer} &
\multicolumn{4}{c|}{ \makecell{whole-body \\ controller}} &
\multicolumn{2}{c|}{ \makecell{low-level \\ joint control}} \\
\cline{3-26}
                     &
                     &
                     \multicolumn{4}{c!{\color{black}\vrule}}{ \makecell{ human motion \\ measurement} }    &
                     \multicolumn{7}{c!{\color{black}\vrule}}{feedback} & 
                     \multirow{3}{*}{\rotatebox[]{-90}{\makecell{~GUI-based planner}}} & 
                     \multicolumn{3}{c!{\color{black}\vrule}}{\makecell{retargeting \& motion \\ generation}} & 
                     \multirow{3}{*}{\rotatebox[]{-90}{\makecell{~ZMP}}} &
                     \multirow{3}{*}{\rotatebox[]{-90}{\makecell{~DCM-ZMP}}} & 
                     \multirow{3}{*}{\rotatebox[]{-90}{\makecell{~contact wrench, \\  force distribution }}} &
                     \multirow{3}{*}{\rotatebox[]{-90}{\makecell{~IK/velocity IK}}} &
                     \multirow{3}{*}{\rotatebox[]{-90}{\makecell{~inverse dynamics}}} & 
                     \multirow{3}{*}{\rotatebox[]{-90}{\makecell{~momentum-based}}} & 
                     \multirow{3}{*}{\rotatebox[]{-90}{\makecell{~QP method}}} & 
                     \multirow{3}{0.6cm}{\rotatebox[]{-90}{\makecell{~joint position}}} & 
                     \multirow{3}{0.6cm}{\rotatebox[]{-90}{\makecell{~joint torque}}} \\ 
\cline{3-13}\cline{15-17}
                     &
                     &
                     \multirow{2}{*}{\rotatebox[]{-90}{\makecell{~motion-capture \\ suit}}} &
                     \multirow{2}{*}{\rotatebox[]{-90}{\makecell{ ~optical-tracking}}} &
                     \multirow{2}{*}{\rotatebox[]{-90}{\makecell{ ~~~~~exoskeleton}}} &
                     \multirow{2}{*}{\rotatebox[]{-90}{\makecell{ mouse,keyboard,\\ \,joystick,treadmill~}}} &
                     \multicolumn{2}{c!{\color{black}\vrule}}{visual} &
                     \multicolumn{5}{c!{\color{black}\vrule}}{haptic}    &
                     &
                     \multirow{2}{*}{\rotatebox[]{-90}{\makecell{~footstep  motion \\ generation}}} &
                     \multirow{2}{*}{\rotatebox[]{-90}{\makecell{~upper-body \\ retargeting}}} &
                     \multirow{2}{*}{\rotatebox[]{-90}{\makecell{~whole-body \\ retargeting}}} &
                      &
                      &
                      &
                      &
                      &
                      &
                      &
                      &
                      \\
\cline{7-13}

                     &
                     &
                     & 
                     &   
                     &       
                     & 
                     \rotatebox[]{-90}{\makecell{ ~mono display, \\GUI~}} &
                     \rotatebox[]{-90}{\makecell{ ~Stereo, \\ AR/VR headset~}} &
                     \rotatebox[]{-90}{\makecell{ ~whole-body \\ exoskeleton}} & 
                     \rotatebox[]{-90}{\makecell{ ~dual-arm \\exoskeleton}} &
                     \rotatebox[]{-90}{\makecell{ ~glove}} & 
                     \rotatebox[]{-90}{\makecell{ ~balance}} & 
                     \rotatebox[]{-90}{\makecell{ ~vibrotactile}} & 
                     &           
                     &              
                     &                 
                     &                 
                     &              
                     &              
                     &                 
                     &          
                     &   
                     &
                     &            
                     &            
                     \\  
\hline\hline

\rowcolor{Gray}
\cite{darvish2019}  &  iCub Genova04 & \checkmark  & ~ & ~ & \checkmark  & ~ & \checkmark & ~  & ~ & ~ & ~ & ~ 
& ~           & \checkmark & \checkmark & \checkmark         & ~ & \checkmark & ~         & \checkmark & ~ & \checkmark & \checkmark         & \checkmark & \checkmark \\

\cite{dafarra2022icub3}  &  iCub3 &           \checkmark  & \checkmark & ~ & \checkmark         & ~ & \checkmark         & ~  & ~ & \checkmark & ~ & \checkmark                  & ~           & \checkmark & \checkmark & ~         & ~ & \checkmark & ~         & \checkmark & ~ & ~ & \checkmark         & \checkmark & ~ \\

\rowcolor{Gray}
\cite{Cisneros2022Team}  &  HRP-4CR &           \checkmark  & ~ & ~ & \checkmark         & ~ & \checkmark         & ~  & ~ & ~ & ~ & \checkmark                  & ~          & \checkmark & \checkmark & ~         & ~ & \checkmark & ~         & ~ & ~ & ~ & \checkmark         & \checkmark & ~ \\

\cite{elobaid2018a}  &iCub Genova04 & ~ & \checkmark & ~ & \checkmark & ~ & \checkmark & ~ & ~ & ~ & ~ & ~ 
& ~           & \checkmark & \checkmark & ~         & ~ & \checkmark & ~         & \checkmark & ~ & ~ & \checkmark         & \checkmark & ~ \\

\rowcolor{Gray}
\cite{jorgensen2019}  & Valkyrie & ~ & ~ & ~ & \checkmark & \checkmark & ~ & ~ & ~ & ~ & ~ & ~ 
& \checkmark          & \checkmark & ~ & ~         & ~ & ~ & \checkmark         & ~ & \checkmark & \checkmark & \checkmark         & ~ & \checkmark \\

\cite{johnson2017team}  & Atlas & ~ & ~ & ~ & \checkmark & \checkmark & ~ & ~ & ~ & ~ & ~ & ~
& ~          & \checkmark & ~ & ~         & ~ & ~ & \checkmark         & ~ & \checkmark & \checkmark & \checkmark         & ~ & \checkmark \\

\rowcolor{Gray}
\cite{chagas2021humanoid}  &  DRC-Hubo &           ~  & \checkmark & ~ & \checkmark         & ~ & \checkmark         & ~  & ~ & ~ & ~ & ~                  & ~          & \checkmark & \checkmark & ~         & \checkmark & ~ & ~         & \checkmark & ~ & ~ & ~         & \checkmark & ~ \\

\cite{penco2019}  & iCub Nancy01 & \checkmark & ~ & ~ & \checkmark & ~ & \checkmark & ~ & ~ & ~ & ~ & ~
& ~          & ~ & ~ & \checkmark         & \checkmark & ~ & ~         & \checkmark & ~ & ~ & \checkmark         & \checkmark & ~ \\

\rowcolor{Gray}
\cite{zucker2015} & DRC-Hubo Beta & ~ & ~ & ~ & \checkmark & \checkmark & ~ & ~ & ~ & ~ & ~ & ~ 
& \checkmark          & \checkmark & ~ & ~         & \checkmark & ~ & ~         & \checkmark & ~ & ~ & ~         & ~ & \checkmark \\

\cite{cisneros2016}  & HRP-2KAI & ~ & ~ & ~ & \checkmark & \checkmark & ~ & ~ & ~ & ~ & ~ & ~ 
& \checkmark           & ~ & ~ & ~         & ~ & \checkmark & \checkmark         & \checkmark & ~ & ~ & ~         & \checkmark & ~ \\

\rowcolor{Gray}
\cite{ishiguro2020bilateral}  & JAXON & ~ & ~ & \checkmark & ~ & ~ & \checkmark & \checkmark & ~ & ~ & ~ & ~
& ~        & ~ & ~ & \checkmark         & ~ & \checkmark & ~         & \checkmark & ~ & ~ & ~         & \checkmark & ~ \\

\cite{ramos2018}  & little HERMES & ~ & ~ & \checkmark & ~  & ~ & \checkmark & \checkmark & ~ & ~ & \checkmark & ~ 
& ~           & ~ & ~ & ~         & ~ & ~ & \checkmark         & ~ & ~ & \checkmark & ~         & ~ & \checkmark \\

\rowcolor{Gray}
\cite{kim2013}  & MAHRU & \checkmark & ~ & ~ & ~ & ~ & \checkmark & ~ & ~ & ~ & ~ & ~ 
& ~           & \checkmark & \checkmark & ~         & ~ & \checkmark & ~         & \checkmark & ~ & ~ & ~         & \checkmark & ~ \\

\cite{schwarz2021nimbro}  &  NimbRo Avatar &           ~  & ~ & \checkmark & \checkmark         & ~ & \checkmark         & ~  & \checkmark & \checkmark & ~ & ~                  & ~          & ~ & \checkmark & ~         & ~ & ~ & ~         & ~ & ~ & ~ & ~         & ~ & \checkmark \\

\rowcolor{Gray}
\cite{eramos2015}  & HRP-2  & ~ & \checkmark & ~ & ~ & ~ & ~ & ~ & ~ & ~ & ~ & ~
& ~           & \checkmark &  ~ & \checkmark         & \checkmark & ~ & ~         & ~ & \checkmark & ~ & \checkmark         & ~ & ~ \\

\cite{ishiguro2018} & JAXON & ~ & \checkmark & ~ & ~ & ~ & \checkmark & ~ & ~ & ~ & ~ & ~
& ~          & ~ & ~ & \checkmark         & ~ & \checkmark & ~         & \checkmark & ~ & ~ & ~         & \checkmark & ~ \\

\rowcolor{Gray}
\cite{tachi2020telesar} & TELESAR VI & ~  & \checkmark & ~ & ~ & ~ & \checkmark & ~ & ~ & \checkmark & ~ & ~         & ~       & ~ & ~ & \checkmark          & ~ & ~ & ~      & \checkmark & ~ & ~ & ~ & \checkmark & ~  \\

\cite{hu2014}  & TORO & \checkmark & ~ & ~ & ~ & ~ & ~ & ~ & ~ & ~ & ~ & ~ 
  & ~         & ~ & ~ & \checkmark         & ~ & \checkmark & ~         & \checkmark & ~ & ~ & \checkmark         & \checkmark & ~ \\

\rowcolor{Gray}
\cite{abi2018}  & TORO & ~ & ~ & \checkmark & ~ & ~ & ~ & ~ & \checkmark & ~ & ~ & ~
& ~         & ~ & ~ & ~         & ~ & ~ & \checkmark         &  ~ & ~ & \checkmark & \checkmark         & ~ & \checkmark \\

\cite{brygo2014b}  & COMAN & ~ & \checkmark & ~ & ~ & ~ & ~ & ~ & ~ & ~ & \checkmark & \checkmark 
& ~          & ~ & ~ & ~         & ~ & ~ & ~         & \checkmark & ~ & ~ & ~         & \checkmark & ~ \\

\rowcolor{Gray}
\cite{difava2016}  & HRP-4 & \checkmark & ~ & ~ & ~ & ~ & ~ & ~ & ~ & ~ & ~ & ~
& ~           & ~ & ~ & \checkmark         & ~ & ~ & ~         & ~ & \checkmark & ~ & \checkmark         & \checkmark & ~ \\

\cite{peternel2013}  & Fujitsu HOAP-3 & ~ & \checkmark & ~ & ~ & ~ & ~ & ~ & ~ & ~ & \checkmark & ~
& ~           & ~ & ~ & ~         & ~ & ~ & ~         & ~ & ~ & ~ & ~         & \checkmark & ~ \\

\hline
\end{tabular}
\arrayrulecolor{black}

\label{tab:teleoperation}
\end{table*}

\subsubsection{Human physiological measurements}
Among the different sensors available to measure human physiological activities, we briefly describe those that have been mostly used in teleoperation and robotics literature. 
An electromyography (EMG) sensor provides a measure of the muscle activity, i.e., contraction, in response to the neural stimulation action potential \cite{staudenmann2010methodological}. It works by measuring the difference between the electrical potential generated in the muscle fibres by employing two or more electrodes. There are two types of EMG sensors, the surface EMG and the intramuscular EMG. The former records the muscle activity from above the skin (therefore, noninvasive), while the latter measures the muscle activity by inserting needle electrodes into the muscle (intrusive). The main problem with EMG sensors, especially the surface one, is the low signal to noise ratio, which is the main barrier for a desirable performance \cite{chowdhury2013surface}. The EMG signals are used in the literature for teleoperating a robot or a prosthesis,  
for estimation of the human effort and muscle forces \cite{staudenmann2010methodological}, or for estimation of the muscle stiffness. The EMG signals can anticipate human motions by measuring the muscle activities within a few milliseconds in advance of force generation; this could be exploited to anticipate the human operator's motion, enhancing the teleoperation. 

Electroencephalography (EEG) sensors can be employed to identify the user mental state. They are most widely used in non-invasive brain-machine interface (BMI) and they monitor the brainwaves resulting from the neural activity of the brain.
The measurement is done by placing several electrodes on the scalp and measuring the small electrical signal fluctuations \cite{niedermeyer2005electroencephalography}.
Other sensors that could be employed in a telexistence scenario for an advanced estimation of the human psycho-physiological state include the Heart Rate Monitor sensors, which estimate the maximal uptake of the oxygen and the heart rate variability \cite{achten2003heart}, useful to measure the user's fatigue, Electro-Oculography (EOG) sensors or video-based eye-trackers, which estimate of gaze position based on the pupil or iris position \cite{sugano2015self}, important for the visual feedback given by Virtual Reality (VR) or Augmented Reality (AR) goggles; and capacitive thin-film humidity sensors, which measure the humidity of the gas-flow of the human skin \cite{ohhashi1998human}, a good indicator of the human emotional stimuli and stress level. 

\subsection{Feedback Interfaces: Robot to Human}
\label{sec:FeedbackInterfaceDevices}
A crucial point in robot teleoperation is to sufficiently inform the human operator of the states of the robot and its work site, so that she/he can comprehend the situation or feel physically present at the site, producing effective robot behaviors.
TABLE \ref{tab:teleoperation} summarizes the different feedback devices that have been adopted for humanoid teleoperation.

\subsubsection{Visual feedback}
A conventional way to provide situation awareness to the human operator is through visual feedback. Visual information allows the user to localize themselves and other humans or objects in the remote environment.
Graphical User Interfaces (GUIs) were widely used by the teams participating at the DRC to remotely pilot their robots through the competition tasks~\cite{johnson2017team,cisneros2016,zucker2015}. 
Not only the information coming from the RGB cameras of the robot but also other information such as depth, LIDAR, RADAR, and thermographic maps of the remote environment was displayed to the user.

An alternative way to give visual feedback to the human operator is through VR headsets, connected to the robot cameras.
Although this has been demonstrated to be effective in several robotic experiments \cite{kim2013,ishiguro2018,darvish2019,penco2019}, during locomotion the users often suffer from
motion sickness, since the images from the robot cameras are not stabilized, while the images perceived by the human eyes are automatically stabilized on the retina thanks to compensatory eye reflexes.
This aspect could be improved by adopting digital image stabilization techniques or AR  \cite{2010ryueyes}.
Another issue concerning the visual feedback is related to the limited bandwidth of the communication link, which can delay the stream of information.
Also, human reaction time to visual input is inherently slow ($\approx$250-300 ms), so a higher delay in the stream of information can be perceived by the user and further aggravate the motion sickness. If also haptic feedback is streamed to the operator, then even lower delays can be disturbing. In fact, human reaction to haptic information is much faster (around 100-150ms).

\subsubsection{Haptic feedback}
Visual feedback is not often sufficient for many real-world applications, especially those involving power manipulation (with high forces) or interaction with other human subjects, where the dynamics of the robot, the contact forces with the environment, and the human-robot interaction forces are of crucial importance. In such scenarios also the haptic feedback is required to exploit the human operator's motor skills in order to augment the robot performance. 

There are different technologies available in the literature to provide haptic feedback to the human. Force feedback, tactile and vibro-tactile feedback are the most used in teleoperation scenarios.
The interface providing kinesthetic force feedback can be similar to an exoskeleton \cite{Wang2015} or can be cable driven. 
The latter only provides a tension force feedback, while the former provides the force feedback on different directions. 
Dual-arm exoskeletons have been proposed to guide the teleoperated robot during manipulation tasks while receiving haptic feedback through the same actuated exoskeleton arms \cite{tachitelexistence, abi2018} and recently also a whole-body exoskeleton cockpit has been proposed to teleoperate the JAXON humanoid robot during heavy manipulation tasks and stepping on uneven terrains \cite{ishiguro2020bilateral}, getting a force feedback on the whole limbs.

To convey the sense of touch, tactile displays have been adopted in the literature \cite{wagner2004design}.
These can also provide temperature feedback to the user.
Other employed haptic feedbacks are vibrotactile and air pressure ones, used as a sensory substitution to transmit senses of touch, texture, forces, suggesting directions, or to catch the attention of the user \cite{brygo2014b}.

All these types of haptic feedback are combined in the telexistence system TELESAR V \cite{tachitelexistence}, which has been developed to provide complete cutaneous sensations to the human operator.
The idea is that different physical stimuli give rise to the same sensation in humans and are perceived as identical. This is due to the the fact that human skin has limited receptors and can perceive only force, vibration, and temperature, which in \cite{hapticcolors} are defined as \textit{haptic primary colors}. It is thus sufficient to combine these \textit{colors} in order to reproduce any cutaneous sensation without actually touching the real object.

\subsubsection{Balance feedback}
Haptic feedback can also be used to transmit to the operator a sense of the robot's balance.
The idea behind the balance feedback is to transfer to the operator the information about the \textit{effect of disturbances} over the robot dynamics or stability instead of directly mapping to the human the disturbance forces applied to the robot. 
In \cite{brygo2014b}, Brygo \textit{et al.} proposed to provide the feedback of the robot’s balance state by means of a vibro-tactile belt.
Also, Peternel and Babic~\cite{peternel2013} proposed a cable-driven haptic interface that maps the state of the robot’s balance to the human demonstrator.
Alternatively, Abi-Farrajil \textit{et al.}~\cite{abi2018} introduced a task-relevant haptic feedback interface composed by two light-weight robotic arms that receives high-level informative haptic cues informing the user about the impact of her/his potential actions on the robot’s balance. 
These studies do not investigate the case of dynamic behaviors, but are rather limited to double support scenarios.
In~\cite{ramos2018} instead, simultaneous stepping is enabled via bilateral coupling between the human operator, wearing a Balance Feedback Interface (BFI), and the robot. The BFI is composed of an passive exoskeleton which captures human body posture and a parallel mechanism which applies feedback forces to the operator's torso.
\subsubsection{Auditory feedback}
Auditory feedback is another means of communication. It is mainly provided to the user through headphones, single or multiple speakers. Auditory information can be used for different purposes: to enable the user to communicate with others in the remote environment through the robot, to increase the user situational awareness, to localize the sound source by using several microphones, or to detect the collision of the robot links with the environment. 
The user and the teleoperated robot may also communicate through the audio channel; e.g., for state transitions.

\subsection{Graphical User Interfaces (GUIs)}
GUIs are used in the literature to provide both feedback to the user and give commands to the robot.
In the DRC, operators were able to supervise the task execution through a task panel, using manual interfaces in case they needed to make corrections. 
The main window consisted of a 2D and 3D visualization environment, the robot’s current and goal states, motion plans, together with other perception sensor data such as hardware driver status~\cite{nakaoka2015task, zucker2015}.
Due to the limited robot cognitive skills, perception tasks were shared among the users and the robot. 

A common approach adopted by the different teams was to guide the robot perception algorithms to fit the object models to the point cloud, used to estimate the 3D pose of objects of interest in the environment. For example, operators were annotating search regions by clicking on displayed camera images or by clicking on 3D positions in the point cloud.
Following that, markers were used to identify the goal pose of the robot arm end effectors \cite{johnson2017team}, the robot configuration, or with a higher autonomy level to define the goal pose of objects for manipulation tasks \cite{nakaoka2015task}.
In these cases, robots tried to find an obstacle-free path (related to Sec. \ref{sec:kinematic-retargeting::high-level}), and show the generated path to the operator for verification prior to execution~\cite{nakaoka2015task}.
Throughout this process, the robots' lower-body teleoperation was treated differently. When the robot's desired base/CoM goal or footsteps were marked by the user, an obstacle-free path (Sec.~\ref{sec:kinematic-retargeting::high-level}) and footsteps trajectories were automatically generated (Sec.~\ref{sec:kinematic-retargeting::low-level}). In this process, footsteps were adjusted to uneven terrain given the estimated height-field data~\cite{nakaoka2015task}.

GUIs have also been used to command frequent high-level tasks to the robot by encoding them as state machines or as task sequences~\cite{nakaoka2015task}.
In DRC open-source software tools, such as RViz and Choreonoid, were commonly used~\cite{zucker2015, nakaoka2015task}.
Custom functionalities were added to them using software plugins. Today, many of these functionalities can be integrated with VR and AR
devices.

\section{Humanoid Robot Retargeting and Control}
\label{sec:RobotControl}

\begin{figure*}[!t]
   \setlength\belowcaptionskip{-0.75\baselineskip}
\begin{minipage}{0.72\textwidth}
\centering
\includegraphics[width=0.99\textwidth]{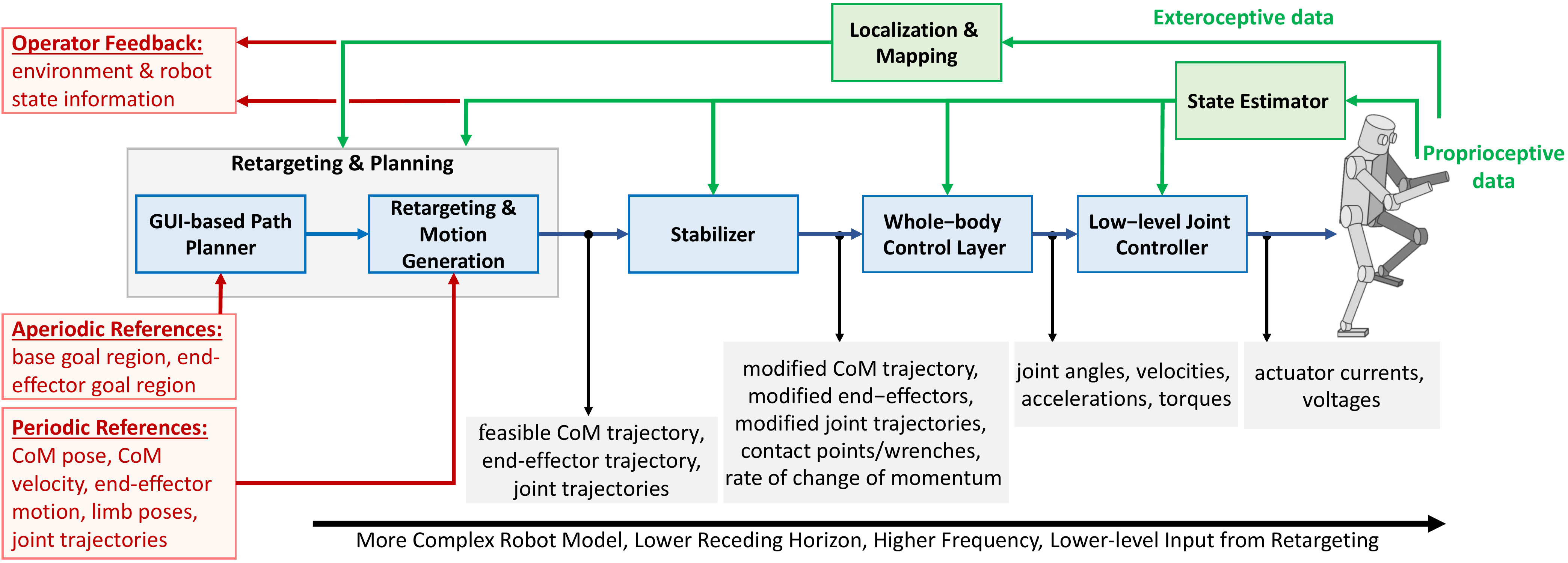}
\end{minipage}
\begin{minipage}{0.27\textwidth}
\caption{Flowchart of a humanoid robot retargeting, planning, and control architecture (red color: references/feedback of human; blue color: retargeting, motion generation, and control; green color: perception and estimation).}
\label{fig:retargeting_controller_architecture}
\end{minipage}
\end{figure*}

This section describes the retargeting and control techniques for unilateral teleoperation of humanoid robots.
We can define the retargeting and control as a mapping $\mathbb{H}: A \to A^{\prime}$ where $A$ is the domain of the perceived human actions (kinematic and dynamic trajectories) and $A^{\prime}$ is the space of robot actions.
In this definition, the mapping principle identifies the robot autonomy and the human authority in the teleoperation scenario.
This mapping should be identified such that it minimizes the difference between the human intent and the robot actions while respecting the constraints in the teleoperation scenario.

Humanoid robot retargeting and control introduces many challenges to teleoperation scenarios, namely due to the nonlinear, hybrid, and underactuated dynamics of humanoids with high degrees of freedom, as well as imprecise robot dynamical model and control, and partially-known environment dynamics.
All these challenges together with the fact that the robot retargeting and controller blocks (in Fig.~\ref{fig:retargeting_controller_architecture}) usually run online, make the problem even more complicated.
To overcome the introduced challenges, model-based optimal control architectures are the foremost technique used in the literature.
During DRC, most of the finalist architectures converged to a similar design \cite{feng2015optimization, johnson2017team} where the robot trajectories and 
stability are achieved by retargeting and control blocks connected in cascade.
This architecture is conceptually demonstrated in Fig.~\ref{fig:retargeting_controller_architecture}, where its blocks are characterized in the following sections.
This architecture needs a humanoid model (Sec.~\ref{sec:Modeling}), human references coming from teleoperation devices, and the robot environment information.
Finally, challenges unique to humanoid robot retargeting, control, and possible future directions are discussed in Sec.~\ref{sec:RetargetingControlChallengesFutureDirections}.

\subsection{Modeling}
\label{sec:Modeling}
\subsubsection{Notations and complete humanoid robot models}
Most humans and humanoid robots are modeled as multi-body mechanical systems with $n+1$ rigid bodies, called links, connected with $n$ joints, each with one degree of freedom.
The configuration of a humanoid robot with $n$ joints depends on the robot shape, i.e., joint angles $\bm{s}\in \mathbb{R}^n$, the position $^{\mathcal{I}}\bm{p}_{\mathcal{B}}\in \mathbb{R}^3$, and orientation $^{\mathcal{I}}{\bm{R}}_{\mathcal{B}}\in SO(3)$ of the floating base (usually associated with the pelvis, ${\mathcal{B}}$) relative to the inertial or world frame $\mathcal{I}$\footnote{ With the abuse of notation, we will drop $\mathcal{I}$ in formulas for simplicity.}. The robot configuration is indicated by $\bm{q} = (^{\mathcal{I}}\bm{p}_{\mathcal{B}}, ^{\mathcal{I}}{\bm{R}}_{\mathcal{B}}, \bm{s})$.
The pose of a frame attached to a robot link $\mathcal{A}$ is computed via $(^{\mathcal{I}}\bm{p}_{\mathcal{A}}, ^{\mathcal{I}}\bm{R}_{\mathcal{A}}) = \bm{\mathcal{H}}_A(\bm{q})$, where   $\bm{\mathcal{H}}_A(\cdot)$ is the geometrical forward kinematics.
The velocity of the model is summarized in the 
vector $\bm{\nu}=(^{\mathcal{I}}\dot{\bm{p}}_{\mathcal{B}},^{\mathcal{I}}{\bm{\omega}}_{\mathcal{B}}, \dot{\bm{s}}) \in \mathbb{R}^{n+6}$, where $^{\mathcal{I}}\dot{\bm{p}}_{\mathcal{B}}$,   $^{\mathcal{I}}{\bm{\omega}}_{\mathcal{B}}$ are the base linear and rotational (angular) velocity of the base frame, and $\dot{\bm{s}}$ is the joints velocity vector of the robot.
The velocity of a frame $\mathcal{A}$ attached to the robot, i.e., $ ^{\mathcal{I}}\bm{v}_{\mathcal{A}}= (^{\mathcal{I}}\dot{\bm{p}}_{\mathcal{A}},^{\mathcal{I}}{\bm{\omega}}_{\mathcal{A}})$, is computed by \textit{Jacobian} of $\mathcal{A}$ with the linear and angular parts, therefore $^{\mathcal{A}}\bm{v}_{\mathcal{I}}= \bm{\mathcal{J}}_A(\bm{q}) \bm{\nu}$.
Finally, the $n+6$ robot dynamics equations, with all $n_c$ contact forces applied on the robot, are described by~\cite{cisneros2020inverse}:
\begin{equation}
\bm{M}(\bm{q}) \dot{\bm{\nu}} + \bm{C}(\bm{q},\bm{\nu})\bm{\nu} + \bm{g}(\bm{q}) = \bm{B}\bm{\tau} + \sum_{k=1}^{n_c}\bm{\mathcal{J}}_{k}^{T}(\bm{q})\bm{f}^{c}_{k},
\label{eq:dyn}
\end{equation}
where $\bm{M}(\bm{q})$ is the symmetric positive definite inertia matrix of the robot, $\bm{C}(\bm{q},\bm{\nu})$ is the vector of Coriolis and centrifugal terms, $\bm{g}(\bm{q})$ is the vector of gravitational terms, $\bm{B}=(\bm{0}_{n\times{6}},\bm{1}_n)^T$ is a selector matrix, $\bm{\tau}$ is the vector of actuator joint torques, and $\bm{f}^{c}_{k}$ is the vector of the $k$-th contact wrenches acting on the robot.
More information about the humanoid robot modeling can be found in \cite{sugihara2020survey}. 

\subsubsection{Simplified humanoid robot models}
Various simplified models are proposed in the literature in order to extract intuitive control heuristics and for real-time lower order motion planning.
The most well-known approximation of humanoid dynamics is the inverted pendulum model \cite{kajita20013d}.
In this model, the support foot
is connected through a variable-length link to the robot center of mass (CoM). Assuming a constant height for the inverted pendulum \cite{kajita20013d}, one can derive the equation of motion of the \textit{Linear Inverted Pendulum Model} (LIPM) by:
\begin{equation}
\ddot{\bm{x}} = \frac{g}{z_0}({\bm{x} - \bm{x}_{b}}),
\label{eq:lipm}
\end{equation}
where $g$ is the gravitational acceleration constant, $z_0$ is the constant height of the CoM, $\bm{x} \in \mathbb{R}^2$ is the CoM coordinate vector, and $\bm{x}_{b} \in \mathbb{R}^2$ is the base of the LIPM coordinate vector.

The base of the LIPM is often assumed to be the Zero Moment Point (ZMP) (equivalent to the center of pressure, CoP) of the humanoid robot \cite{vukobratovic1972stability}.
The LIPM dynamics can be divided into stable and unstable modes, where the unstable mode is referred as (instantaneous) Capture Point \cite{pratt2006},  or Divergent Component of Motion (DCM) \cite{takenaka2009real, englsberger2015}
in the literature.
The DCM dynamics is characterised by a first-order system as:
\begin{equation}
{\bm{\xi}} = \bm{x} + b{\bm{\dot{x}}},
\label{eq:dcmCOM}
\end{equation}
where $\bm{\xi}$ and $b = \sqrt{\frac{z_0}{g}}$ are the DCM variable and the time constant. Equation \ref{eq:dcmCOM} shows that the CoM follows the DCM.
Differentiating Eq.~\eqref{eq:dcmCOM} and replacing into Eq.~\eqref{eq:lipm} results in:
\begin{equation}
\dot{\bm{\xi}} = \frac{1}{b} (\bm{\xi}- \bm{x}_b).
\label{eq:dcm}
\end{equation}
Equations \eqref{eq:dcmCOM} and \eqref{eq:dcm} together represent the LIPM dynamics.

\subsection{Retargeting \& Planning}
\label{sec:KinematicRetargeting}

The goal of this block in Fig.~\ref{fig:retargeting_controller_architecture} is to morph the human commands or measurements coming from the teleoperation devices into robot references.
It comprises the reference motions for the robot links as well as the robot locomotion references, i.e., alternate footstep locations, timings, and allocating a given footstep to the left or right foot to follow the user's commands.
The input to this block may vary according to the task and system requirements, and consequently the design choice and retargeting policy differ.
Besides, the retargeting policy can even be determined online as a classification or regression problem using the human speech and psychophysiological state. For example in  \cite{Singh2018Physiologically, Dragan2012}, the authors developed attentive systems for real-time adaptation of the retargeting policy and the robot autonomy level.
As we go from the left to the right in Fig.~\ref{fig:retargeting_controller_architecture}, the level of automation and the input frequency increase, and higher communication bandwidth with lower delays is required.
The approaches presented in this part are model-based, while learning-based techniques for retargeting will be discussed later in Sec.~\ref{sec:RetargetingControlChallengesFutureDirections}.

Teleoperation devices may provide different types of input to retargeting and planning. The inputs can be
\textit{i}) the desired goal pose (region) in the workspace for the base and the end-effectors using GUIs (high-level),
\textit{ii}) the CoM velocity, the base rotational velocity, the end-effector motion, the user's desired footstep contacts, or whole-body motion (low-level).
While at a low-level, the user is in charge of the obstacle avoidance, the planning and retargeting at the high-level deals with finding the path leading to the desired goal.
The output of the high-level goes to the lower level in order to finally compute the robot reference joint angles, footstep contacts and timings as the output of the  block.
We structured the rest of this subsection according to the input category.

\subsubsection{GUI-based Path Planner}
\label{sec:kinematic-retargeting::high-level}
When the user provides the goal region of the robot base or arms through GUIs, the humanoid robot not only should plan its footsteps or arm motion but also should find a \textit{feasible} path for reaching the goal, if any exists. Related to the footsteps, contrary to wheeled mobile robots, a feasible path here refers to an obstacle-free path or one where the humanoid robot can traverse the obstacles by stepping over.
Primary approaches for solving the high-level path planning are search-based methods and reactive methods.
The first step toward finding a feasible path, i.e., regions where the robot can move, is to perceive the workspace by means of the robot perception system. Following the identification of the workspace, a feasible path is planned.

Search-based algorithms try to find a path from the starting point to the goal region by searching a graph. The graph can be made using a grid map of the environment or by a random sampling of the environment.
Some of the methods used in the literature to perform footstep planning are A* \cite{griffin2019footstep}, D* Lite \cite{garimort2011humanoid}, RRT variations \cite{perrin2011fast}, and dynamic programming techniques \cite{kuffner2001footstep}.
The \textit{completeness}, \textit{global optimality}, and the ability of real-time \textit{replanning} of the path in dynamic environments are the important features of these search-based algorithms when selecting a proper method.
These algorithms are not very efficient for real-time execution when an exhaustive search is done, hence, to enhance the efficiency a \textit{heuristic} is chosen in order to prune the search space and perform a greedy search.

Reactive methods for high-level kinematic retargeting and path planning problems can be addressed as an optimization problem or as a dynamical system.
In \cite{herdt2010online} the problem has been tackled with a Model Predictive Control (MPC) approach. It allows finding the foot poses as a continuous decision-making problem by formulating it as a Quadratic Programming (QP) optimization problem.
However, when the end-effector rotation or the obstacle-avoidance is added to the problem, the optimization problem becomes non-convex, therefore, there is no guarantee on completeness and global optimality \cite{deits2014footstep}.
However, although there are approaches to relax the non-convex optimization problems, their computational complexity is still a challenge \cite{deits2014footstep}.
Moreover, the problem of path planning and obstacle avoidance for a humanoid robot can be viewed as a simplified dynamical system control approach, for example, by using potential fields \cite{fakoor2015revision}.

\subsubsection{Retargeting \& Motion Generation}
\label{sec:kinematic-retargeting::low-level}
Given the continuous measurements from the teleoperation devices, we can divide the retargeting approaches into three groups: \textit{lower-body footstep motion generation}, \textit{upper-body retargeting}, and \textit{whole-body retargeting}.

\paragraph*{Lower-body footstep motion generation}
The role of this block is to plan the footstep motion and find the sequence of foot locations and timings given the CoM position or velocity, and the floating base orientation or angular velocity provided by teleoperation devices.
One possible approach to address this problem is based on the instantaneous capture point \cite{pratt2006}.
Given the reference CoM position and velocity, one can compute the next 
foot contact point using the capture point relation in \eqref{eq:dcmCOM}.
In simple cases, the contact sequences can be preliminarily identified by the user for example by a finite state machine, and the desired footstep locations are modified to the left or right side of the capture point according to the nature of the foot contact (left or right).
Another way to find the sequence of footsteps is to formulate an optimization problem, where the cost function is decided based on commonsense heuristics.
The footstep timing can be found from the CoM velocity such that the total gait cycle duration corresponds to the average CoM velocity and gait length.
This approach uses minimal information to generate footstep motion and does not enforce the user and the robot motion similarity.

\paragraph*{Upper-body retargeting}
In this method, the retargeting is done either in task space or configuration space. 
In the task space retargeting, the Cartesian pose (or velocity) of some human limbs is mapped to corresponding values for the robot limbs.
Later, the inverse kinematic problem is solved by minimizing a cost function on the basis of the robot model while considering the robot constraints \cite{darvish2019}.
Different authors considered disparate limbs as the target of the mapping. A popular approach is to map the motion of the human wrist to that of the robot end effectors \cite{elobaid2018a, ishiguro2018}.
A commonly used mapping in the literature is to perform an identity map between the rotational motion of the human and the robot, whereas in the case of translational motion a fixed gain (due to differences in the geometry) is used \cite{elobaid2018a}.
A more complicated approach may identify this rotational and translational gain as a function of the human intention and ongoing task.
To enhance the similarity of the robot motion to the human one, the retargeting of the elbow motion with a lower priority in the optimization problem is suggested
in~\cite{Liarokapis2013}.
To overcome the manual morphing problem, \cite{Ayusawa2017} 
 suggested solving an optimization problem to identify geometric parameters of the morphing function.

Configuration space retargeting refers to the mapping in the joint space from human to robot. This morphing is normally used when the human and robot have similar joint orders.
In this technique, human measurements and model are used in an inverse kinematics problem. The joint angles and velocity of the human joints are identified and mapped to the corresponding joints of the robot \cite{Liarokapis2013}.
When the human joint ranges differs from those of a humanoid, the robot constraints should also be applied in the morphing function.

\paragraph*{Whole-body retargeting}
This method measures the whole-body motion of human and kinematically retargets to the robot motion, similar to the upper-body retargeting approaches. Given the environment information, this approach may consider the contact constraints in the retargeting phase. Thus, this approach yields footstep locations, sequences, and timings. Later, outputs of this technique are provided to the \textit{stabilizer}, to deliberate on the feasibility and enhance the robot stability \cite{ishiguro2018}.
In~\cite{penco2019}, authors measured the normalized ground projection of human CoM from an arbitrary foot and retargeted it to the equivalent robot CoM ground projection, on a line connecting the two robot feet. This approach can be extended toward retargeting the heel-to-toe motion (orthogonal to the feet line) and multi-contact scenarios~\cite{otani2017}.

\subsection{Stabilizer}
\label{sec:Stabilizer}
The main goal of the \textit{stabilizer} is to implement a control policy that \textit{dynamically adapt} input references to enhance the stability and balance of the centroidal dynamics of the robot. Because of the complexity of robot dynamics, classical approaches are limited to examine the stability of a closed-loop control system.
Therefore, other insights such as ZMP criteria and DCM dynamics are tailored in order to evaluate how far the robot is about to fall \cite{koolen2012}.
The \textit{stabilizer} gets inputs from the \textit{retargeting \& planning} level, as shown in Fig.~\ref{fig:retargeting_controller_architecture}.
However, these reference trajectories may destabilize robot's behavior, therefore the \textit{stabilizer} adapts those references based on different criteria.
Accordingly, the output of the \textit{stabilizer} are references for the CoM position, the end-effector poses, joint angles, contact points, contact wrenches, and/or the rate of change of the momenta.
Next, we provide an overview of stabilization approaches according to different criteria adopted so far in the literature.

\begin{figure}[!t]
 \setlength\belowcaptionskip{-1.1\baselineskip}

\centering

\includegraphics[width=0.9\columnwidth]{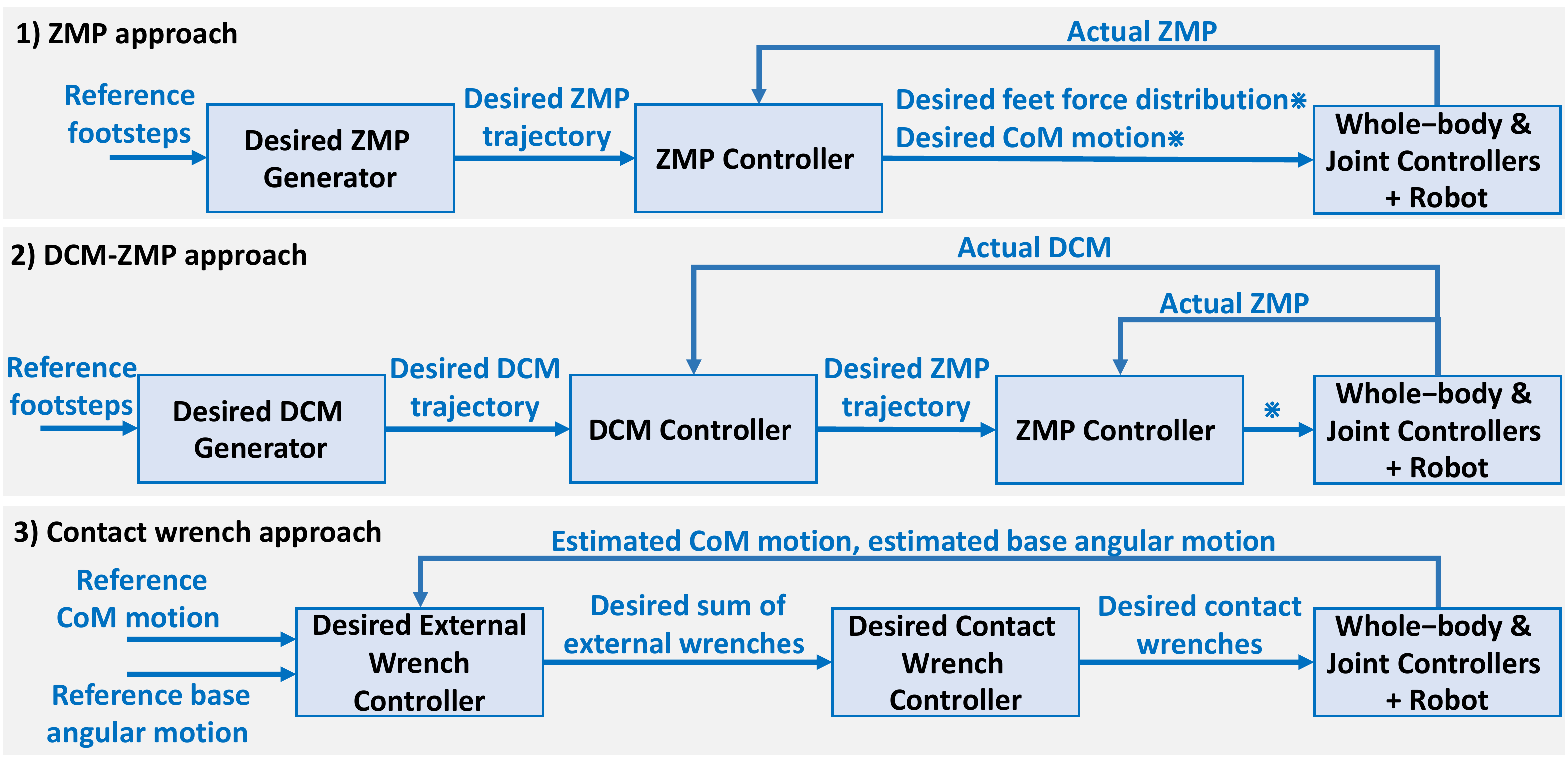}

\caption{Different stability criteria approaches.
}
\label{fig:StabilizerControllers}
\end{figure} 

\subsubsection{ZMP approach}
This approach is based on the idea that the robot's ZMP should remain inside of the support polygon of the base \cite{vukobratovic1972stability, koolen2012}, as shown in Fig.~\ref{fig:StabilizerControllers}.
Given the footstep trajectories provided by the kinematic retargeting, this approach computes the desired ZMP.
While walking, in single support the desired ZMP remains in the middle of the support polygon (e.g., left foot), and during double support the desired ZMP moves smoothly from the previous support leg to the middle of the new support polygon (e.g., from the left foot to the right foot).
Following that, a ZMP controller makes sure that the real ZMP tracks the desired ZMP trajectory. 
However, the ZMP controller introduces several limitations.
If the robot is subject to high disturbances, it does not adapt online to footstep locations to avoid the robot from falling.
Moreover, this approach hardly extends to non-flat terrain or when the support foot rotates or slides.

\subsubsection{DCM-ZMP approach}
Another approach that is often employed with the simplified model retargeting and control is the DCM \cite{takenaka2009real}, which can be viewed as an extension of the capture point concept to the three-dimensional case for uneven terrain \cite{englsberger2015}.
Eq.~\eqref{eq:dcm} shows that the DCM dynamics is unstable, i.e., it has a strictly positive eigenvalue, and using equation of LIPM \eqref{eq:lipm} it can be shown that the CoM dynamics converges to the DCM value.
Therefore, the main goal of this controller is to implement a control law that stabilizes the unstable DCM dynamics \eqref{eq:dcm} as well as the ZMP as mentioned previously.
To stick to the formulation presented in Sec.~\ref{sec:Modeling}, the DCM-ZMP dynamic retargeting is explained with flat floor assumption; however, for the extension of the controller to 3D, one can refer to \cite{englsberger2015}.
The block diagram of the DCM-ZMP stabilizer is also shown in Fig.~\ref{fig:StabilizerControllers}.
DCM-ZMP approach has been deployed for whole-body retargeting in \cite{ishiguro2018}.
In order to enhance the stability of the system, the authors have introduced the predicted support region where the time-delayed DCM of the robot is kept inside that region.

\subsubsection{Contact wrench approach}
This approach finds the desired contact wrenches at each contact point such that it enhances the robot stability. A schematic block diagram of this controller is shown in Fig.~\ref{fig:StabilizerControllers}.
This controller has been initially introduced in \cite{hirukawa2006universal} and later considered by \cite{ott2011} as a stability augmentation criteria when introducing the contact constraints. This approach is similar to the momentum-based controller and will be explained in more detail later. However, to be thorough, we have mentioned it here as well.

\subsection{Whole-Body Control Layer (WBC)}

The \textit{WBC} block in Fig.~\ref{fig:retargeting_controller_architecture} gets as input the retargeted human references corrected by the \textit{stabilizer}, and provides as an output the robot joint angles, velocities, accelerations, and/or the joint torques.
The whole-body control problem can be formalized with different cost functions and be solved as a QP problem or other approaches such as Linear Quadratic Regulator (LQR) \cite{mason2014, kuindersma2016optimization} and MPC \cite{ kelly2017introduction, posa2014direct}.
In the following, different whole-body control approaches are presented.

\subsubsection{Whole-body inverse instantaneous velocity kinematics control}
The problem of inverse instantaneous or velocity kinematics is to find the configuration state velocity vector $\bm{\nu}(t)$ for a given set of task space velocities using the Jacobian relation.
One common approach is to formalize the controller as a constrained  QP problem with inequality and equality constraints.
Conventional solutions of redundant inverse kinematics are founded on the pseudo-inverse of the Jacobian matrix \cite{kanoun2011}.
\subsubsection{Whole-body inverse kinematics control}
The problem of Inverse Kinematics (IK) is to find the configuration space vector $\bm{q}(t)$ given the reference task space poses.
This problem can be sometimes solved analytically for determined robots; however, this solution is not scalable to different architectures and for a redundant humanoid robot (with a high degree of freedom), it is not feasible.
Differently from inverse instantaneous velocity kinematics, to solve the IK problem a nonlinear constrained optimization problem is defined using the geometrical forward kinematics relation.
Solving this problem might be time-consuming and the results can be discontinuous as well.
To solve these challenges, a common approach in the literature is to transform the whole-body IK into a whole-body inverse velocity kinematics problem.

\subsubsection{Whole-body inverse dynamics control}
The Inverse Dynamics (ID) refers to the problem of finding the joint torques of the robot to achieve the desired motion given the robot's constraints such as joint torque limits and feasible contacts.
There are different techniques to solve the ID problem in the literature \cite{righetti2012quadratic}.
To reach the desired end-effector poses and contact wrenches, one can formulate the ID problem as an optimization problem, minimizing the error on metrics such as the motion tasks and contact wrenches. Some of the constraints are dynamics equation of motion \eqref{eq:dyn}, joint angle, velocity, acceleration, or torque limits, collision avoidance constraints, and non-sliding contact constraints.
The details of the constraints can be found in~\cite{difava2016, righetti2012quadratic}.

\subsubsection{Momentum-based control} This controller finds the configuration space acceleration and ground reaction forces such that the robot follows the given desired rate of change of whole-body momentum \cite{orin2013centroidal, cisneros2020inverse}.
According to the Newton-Euler's laws of motion and \eqref{eq:dyn}, the rate of change of centroidal momentum (the whole-body momenta of the humanoid robot about the CoM) is equal to the sum of all external wrenches applied to the robot. Using that, one can write the momentum-based controller in a QP fashion~\cite{koolen2016}.
Eventually, external wrenches and joint accelerations are used with an inverse dynamics algorithm to compute the robot joint torques.

\subsection{Low-level Joint Controller}
One of the main challenges in deploying a humanoid robot controller is the different behavior obtained in simulation and on the real robot of the low-level joint torque tracking.
The low-level joint controller is in charge of generating motor commands to ensure the tracking of the higher-level control inputs.
The input to the joint controller can be the desired joint positions, velocities, torques, or a mixture of them. The output of this controller is the current or voltage to the motors driving the joints.
While the joint position can be directly measured by the encoder sensors, the joint velocity feedback is obtained by differentiating in time the encoder values, hence it can be noisy. Joint torque sensors or whole-body estimation algorithms equipped with distributed joint-torque sensors can estimate the generated joint torques. 
The torque control of the robot joints is more challenging because of the dynamics of the joints which are subject to friction and the mechanical power transmission from the motor to the joint shaft.
On the other hand, compliance control of the joints allows for more robust locomotion and a safer interaction of the robot with the humans and the environment.
For example, ankle joint torque control allows for conforming the foot to the ground in case of small obstacles or a mismatch between the ground slope and its estimation.
However, the compliance control with only joint torques may lead to poor velocity or position tracking, therefore a mixture of them is proposed in the feedback and feed-forward terms of the joint controller in \cite{johnson2017team, kuindersma2016optimization, cisneros2020inverse}.

\subsection{State Estimator, Localization \& Mapping}
These blocks in Fig.~\ref{fig:retargeting_controller_architecture} receive measurements from the robot sensors and estimate the necessary information for other blocks in Fig.~\ref{fig:retargeting_controller_architecture} or to the human as shown in Fig.~\ref{fig:teleoperation_architecture} (for assisted teleoperation). 
A family of well-known model-based estimation techniques commonly used in robotics is the Kalman filters.
The joint encoders are used to estimate the joint angles, velocities, and acceleration.
These joint states accompanied by the robot dynamics model in Eq.~\eqref{eq:dyn} and force/torque sensors can be used to estimate joint torques and external forces~\cite{flacco2016residual}.
To estimate the joint torques, the actuation system model can be learned or identified as well~\cite{hwangbo2019learning}.
If a humanoid robot is equipped with tactile sensors, the external wrenches and its point of application can be estimated likewise~\cite{chavez2018contact}.
To estimate the ground reaction wrenches and ZMP of a humanoid, its feet are normally equipped with force/torque sensors~\cite{kajita2010}.
Moreover, a combination of proprioceptive and exteroceptive sensory information allows a legged robot to estimate better the terrain characteristics and eventually elevate the control robustness~\cite{miki2022learning}.
Finally, calculated joint angles and velocities are used to estimate the robot link poses and velocities through the forward kinematics. Combining those values with the robot link inertia, one can compute the CoM position and velocity~\cite{kuindersma2016optimization}.
Moreover, considering the uncertainties of the link inertia and joint measurement noises, CoP and force/torque sensors are used to estimate CoM in~\cite{atkeson2012state}.

One of the challenges specific to humanoid robots is the estimation of the robot base, and eventually robot localization in an environment.
For the estimation, either proprioceptive sensors (i.e., joint encoders and IMU sensor attached to a robot link) or a combination of proprioceptive and exteroceptive sensors (such as cameras, GPS) are employed.
When only proprioceptive sensors are used, a common approach is to assume that at each time instant at least one of the robot links is in contact with the ground, therefore the frame attached to the contact link has zero velocity and no slippage. With this assumption and taking into account the floating base frame pose in the kinematic chain, one can estimate the floating base velocity given the joint velocity vector, and eventually the base pose by integrating those velocities. However, the error of estimation is propagated over time due to kinematic modeling errors.
To enhance the accuracy and limit the uncertainty of state estimation endowed with the odometry, IMU measurements are fused in the estimation process \cite{rotella2014state, bloesch2013state}.
Nevertheless, exploiting only the proprioceptive sensors does not lead to observability in the yaw axis (parallel to gravity vector) and the absolute position of the robot \cite{bloesch2013state}; thereby exteroceptive sensors can facilitate to overcome this difficulty.
Exteroceptive data such as camera information allows finding feasible regions for the humanoid robot foot locations as well as a map of the environment and obstacles.

\subsection{Challenges \& Future Directions for Retargeting \& Control}
\label{sec:RetargetingControlChallengesFutureDirections}

Humanoid robot teleoperation is a new field, and many challenges to put together whole-body coordinated motion retargeting, planning, stability, and control are not addressed effectively yet.
For example, dividing the retargeting and planning problems of humanoid robot teleoperations appears to be useful but not effective in performing agile teleoperation tasks outside of the lab (in the real world) and in unstructured environment as it is the case for many hazardous environments and disaster response scenarios.
Many of the techniques adopted so far use simplified models of humanoid robots. Although these approaches are computationally efficient, they are limiting due to the adopted simplifying assumptions, e.g., fixed robot CoM height, ignorance of human or robot rotational motion, the existence of at least one contact point with no slippage between the robot and environment.

An alternative approach to solving simultaneously whole-body coordinated retargeting and planning problems is using the MPC technique and
defining them as optimization problems with equality and inequality constraints. However, they are non-linear and non-convex optimization problems with large input and state spaces; therefore, they are computationally demanding, and the optimization may suffer from local minima.
This issue intensified when retargeting the rotational motion and angular momentum of the human to the robot, while biomechanical studies show their importance as one major underlying component of the human-like coordinated motion~\cite{herr2008angular}.
To this goal, an \textit{angular excursion} index has been introduced by \cite{zordan2014control} relating the whole-body orientation, angular velocity, and centroidal angular momentum. When this idea integrates with linear momentum, it enables highly agile motion with considerable rotational behaviors \cite{Kuindersma2020Youtube, zordan2014control}.
However, retargeting the human angular excursion to the robot is an open problem.
A possible trade-off solution between simplified and whole-body motion retargeting and planning approaches is to adopt the centroidal dynamics, the terrain map, and whole-body kinematics of the robot given the human measurements \cite{dai2014whole, Hutter2022Youtube}.
This approach benefits from centroidal dynamics constraints and whole-body kinematics in collision avoidance and reachability computations. However, to adopt MPC in humanoid teleoperation scenarios, major challenges remain to predict future human motion and the difference between the surrounding terrain of human and robot. 

While the MPC approach can be a valid solution for the humanoid robot teleoperation, it is not sufficient when deploying the robot in the real world to perform various tasks in an unstructured and dynamic environment with varying compliance and slippery characteristics.
An approach to resolve these challenges is to introduce new sets of manual heuristics and constraints to the optimization problem for every scenario and environment. However, it is tedious and inefficient to generalize over different situations.
To overcome these drawbacks, data-driven approaches with special attention to the robot dynamics and stability indications have shown promising results in the learning and mobile robot communities.
\cite{lin2019efficient, peng2018deepmimic, xie2019iterative} are some of the successful examples of leveraging neural networks and reinforcement learning (RL) techniques.
In retargeting problems, safety considerations can be explicitly enforced as well \cite{choi2020cross}.
A novel approach blending an unsupervised learning technique with forward kinematics is proposed by \cite{villegas2018neural}. It relies on the cycle consistency principle, i.e., motions retargeted to the avatar should generate the original motions of the human when retargeted back.
The general trend in whole-body planning of robots is that many robots simultaneously learn how to perform agile and dynamic motion robustly in a simulated environment, by incrementally introducing new complex terrain difficulties, dynamic obstacles, and terrain with different parameters. To relax the simulation to real-world gap, \cite{hwangbo2019learning} suggested learning robot actuator dynamics from the real robot and incorporating them with simulated robots.
However, none of those works considered at the same time the whole-body coordinated motion retargeting of a human to a humanoid robot. We speculate that adding a reward term in an RL problem or using transfer learning techniques can impose the similarity of the human and humanoid robot motion during the training phase. While in some cases, human motion can be tracked by the humanoid robot, in other cases, the humanoid robot may perform step adjustments to keep its balance, for example, based on some behavioral selection techniques \cite{Hutter2022Youtube}.

\section{Assisted teleoperation}
\label{sec:AssistedTele}

In many teleoperation scenarios, controlling the robot as explained in the previous section, is not the only viable solution.
In fact, many tasks can achieve higher performances by sharing the autonomy among the human and robot. This section provides more details about these assisted teleoperation strategies.

\subsection{Shared Control}
Delegating robot's full control to the human operator's experience can often limit the efficiency of the teleoperation, resulting in clunky motions, failures, or numerous attempts before being able to accomplish a given task.
This applies particularly to humanoid robots where the operator has to control many aspects at once via teleoperation (e.g., the pose of both hands, the feet location, balance) and can fail very easily without significant robot autonomy being used simultaneously.
In shared-control teleoperation, some robot autonomy is used to assist the user in accomplishing the desired task, potentially making teleoperation easier and more seamless. Generally, the operator's input is modified according to specific metrics by sharing the control authority between the robot and the operator to enhance performance or safety \cite{dragan2013}. 
For example, in \cite{rakita2019}, Rakita \textit{et al.} teleoperated the upper-body of a humanoid using a shared-control approach, providing on-the-fly assistance to help the user complete tasks more easily, enhancing the end-effectors control while performing bi-manipulation of objects. Similarly, in \cite{rahal2019},  Rahal \textit{et al.} designed a shared control approach to assist the human operator by enforcing different nonholonomic-like constraints representative of the cutting kinematics. In other shared-control approaches, the user provides an input $\bm{u}$, which enables the robot to predict human intent, and assist her/him in the task by adjusting the motion or by executing a pre-optimized version of that motion \cite{dragan2013}. Then, a blending policy arbitrates the user input $\bm{u}$ and the enhanced robot motion $\bm{r}$, determining the final reference:
\begin{equation}
\bm{u^*} = (1-\alpha)\bm{u}+\alpha\bm{r},
\label{eq:sharedcontrol}
\end{equation}
where $\alpha$ can be any scalar function. A common choice is the confidence in the prediction of the user intent:
\begin{equation}
\alpha = max(0, 1 - d/D),
\label{eq:alpha}
\end{equation}
where $d$ the distance to the goal and $D$ some threshold past
which the confidence is 0. In this case the closer the robot gets to a predicted goal, the more likely that this goal is the correct one, and the input $\bm{r}$ is preferred over $\bm{u}$.
The prediction of the user intent has also been successfully used to provide haptic guidance through a master device to teleoperate a robot manipulator \cite{Ly2021}, and could be applied to humanoid robots by using exoskeletons as input devices. The haptic information can also be used to enhance the user's comfort during teleoperation \cite{Rahal2020}.

\subsection{Supervised and Safeguarded Teleoperation}
\label{sec:safeguardedteleop}
When full robot autonomy is available for a given task, the operator can simply act as a supervisor. By monitoring the robot, the operator can then identify and react to unexpected problems and intervene in a timely manner by controlling the robot directly to handle “uncovered” situations. This was a common approach in the DRC, where team operators could guide the robot to achieve complex tasks through failure when needed \cite{johnson2017team}.
Similarly, in \cite{dylan2013}, the user monitored multiple robots interacting with passersby in a shopping mall. The robots performed their own speech, gesture, and motion planning autonomously, and the role of the human was only to provide occasional sensor inputs. In rare cases, an operator had to control the robot directly to handle unexpected questions from a customer or to re-plan the robot’s path to avoid unmodeled obstacles.

A mirrored approach can also be adopted when teleoperating robots. For example, in \cite{fong2001}, the operators shared control with a safeguarding system onboard the robot. In benign situations, the
operator had full control of the vehicle's motion, while in hazardous situations, the safeguarder modified or overrode the operator's commands to maintain safety. The safeguarder, therefore, exhibits many characteristics of autonomous systems, such as perception and command generation. In \cite{fong2001}, the safeguarder not only had the function of preventing collision and rollover but also monitoring the system's health (e.g., vehicle power, motor stall). A similar approach could be used for humanoid robots to prevent them from falling or reaching singular configurations while being operated by the user.

\subsection{Bilateral Teleoperation}
\label{sec:bilateral}
Bilateral teleoperation techniques have been widely used in the literature for robot manipulators.
In this approach, not only the robot receives kinodynamic references from the human, but the operator also receives kinesthetic feedback (force, pressure, vibration, etc) from the robot. This feedback informs the operator about the robot's performance reproducing the commanded motion or about external disturbances applied to the machine.
The approaches described next focus on teleoperating the upper- or whole-body, which in the latter case couples altogether the human operator and the robot at upper- and lower-body levels.

\subsubsection{Upper-body bilateral teleoperation}
A simpler strategy that has been adopted on humanoid robots consists in teleoperating the upper-body in a bilateral fashion, while using a separate balancing controller on the lower-body to regulate balance \cite{brygo2014a,peer2008}. To control the upper limbs of the HRP-2, Peer \textit{et al.} \cite{peer2008} adopted a common control scheme for position-controlled robots: the \textit{admittance} controller. Such controllers define the reference through the differential equation of a virtual mass-spring-damper system.
Under time delay, the parameters of the controller should be selected appropriately to guarantee stability of the overall teleoperation system, as done in \cite{evrard2009}
while teleoperating the HRP-2 located in Tsukuba (Japan) from Munich (Germany).

\subsubsection{Whole-body bilateral teleoperation}
\begin{figure}[!t]
  \setlength\belowcaptionskip{-0.7\baselineskip}

    \centering
    \includegraphics[width=2.5in]{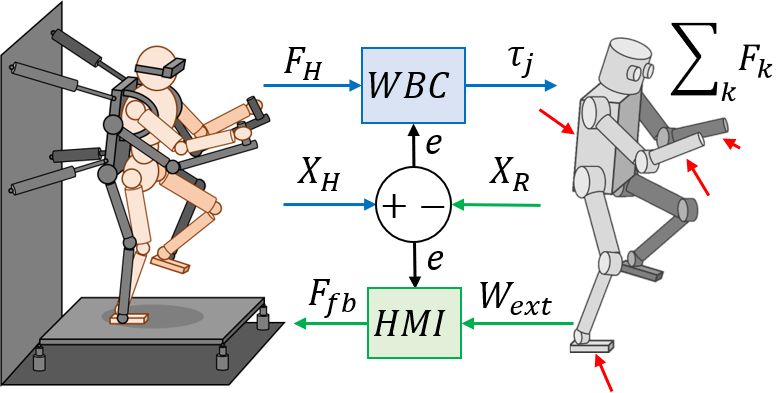}
    \caption{
    General concept for
    whole-body bilateral teleoperation. The robot  WBC computes joint torques $\tau_{j}$ using the reference interaction forces $F_H$ from the operator and the error $e$ between the human state $X_H$ and the robot state $X_R$. The external contact forces $F_k$ applied to the robot create a net wrench $W_{ext}$, which is used by the human-machine interface (HMI) to compute a proportional kinesthetic feedback $F_{fb}$ to the operator. 
    }
    \label{fig:Bilateral_Teleoperation}
\end{figure}

An extension of the conventional idea of bilateral teleoperation is starting to be investigated to dynamically couple human and robot at a whole-body level \cite{ramos2018}. This strategy consists of mapping the whole-body kinematic (joints position, velocitiy, etc) and dynamic (contact forces, joint torques, etc) references from  humans to robots, while providing the operator with feedback regarding the robot's whole-body dynamics, as shown in~\text{Fig.}~\ref{fig:Bilateral_Teleoperation}. However, the naturally unstable dynamics of humanoid robots poses an additional challenge to the whole-body teleoperation: the robot
must balance while reproducing human movement. 

The strategies for whole-body bilateral teleoperation utilize kinesthetic feedback to inform the operator about the robot's dynamics and stability in real-time. The strategy usually focuses on the CoM dynamics, and other condensed information about the robot. For instance, the cable driven feedback interface in \cite{peternel2013} exerts forces on the demonstrator’s waist corresponding to the state of the robot’s CoM. This feedback allows the human to teach the robot how to compliantly interact with the environment. In \cite{Wang2015}, the feedback force applied to the human's torso is proportional to how close the robot is from tipping over. This is estimated by considering the distance between the robot's CoP and the edge of the support polygon. The closer the robot is from tipping over in one direction, the larger the feedback force applied to the operator in the opposite direction. A similar strategy is used in \cite{brygo2014b} by providing discrete vibration levels to the operator using a belt with vibrotactile feedback. In \cite{ishiguro2020bilateral}, the force feedback device TABLIS, a powered exoskeleton, applies forces to the operator's feet to indicate that the robot is stepping onto an obstacle. This enables the operator to control the robot to navigate over objects. Finally, in \cite{ramos2019dynamic} the force feedback is utilized to dynamically synchronize human and machine. The force feedback generates drag (negative feedback) if the robot cannot keep up with the operator's movement, or it speeds up human motion (positive feedback) if the robot moves faster than the operator. The high-level expression for the force feedback is given by:
\begin{equation}
    \bm{f}_{fb} = k_{H}\left[\left( \dot{\bm{x}}_R' - \dot{\bm{x}}_H' \right) +  \bm{f}_{ext}'\right],
\end{equation}
where $\dot{\bm{x}}_i'$ is the dimensionless CoM velocity of the human ($H$) and robot ($R$) \cite{pratt2006}, $\bm{f}_{ext}'$ is dimensionless external force vector applied to the robot, and $k_H$ is a scaling factor proportional to the operator's size and body mass. This strategy enables human and robot to dynamically take simultaneous steps. 

In general, during whole-body bilateral teleoperation, the kinesthetic feedback provided to the operator is proportional to some kinematic or dynamic discrepancy between human and robot with respect to the task at hand and/or the balance regulation. The assisted teleoperation principle arises from the fact that the robot must prioritize between following the human motion command to perform a task and maintain its own bipedal stability. Some strategies shift the balancing authority to the robot's autonomous controller. This means that the robot is responsible for predicting if a given command will jeopardize stability and deciding the best course of action to override the motion. For instance, in \cite{ishiguro2018}, the robot's controller uses stability considerations from the LIP model to modify the CoM trajectory commanded by the human, preventing the operator from destabilizing the robot. This represents a more conservative approach that guarantees stability of the movement of the robot, but prevents the operator from exploring motions that would go beyond the boundaries of the stability metric employed. In contrast, other approaches, such as \cite{Ramos_Humanoids2015}, rely on the human operator to actively regulate the robot's balance. 
In this sense, the robot follows the human's movement with only minor regulation from the robot's autonomous controller; then,
the operator must perceive the robot's destabilization through the kinesthetic feedback and mitigate it by adapting the commanded movement. Although riskier, this strategy frees the operator from abiding to the boundaries of predefined stability metrics, which are challenging to be mathematically defined in the context of legged robots. The goal of this approach is to eventually allow the operator to learn how to cope with the robot's dynamics and to create new motions on the fly. 

\subsection{Impedance Control}
A similar concept to admittance control can be employed in non-bilateral systems to provide the robot the ability to dexterously interact with the remote environment. In \cite{brygo2014a}, Brygo \textit{et al.} implement an autonomous \textit{impedance} controller that regulates the robot’s joints stiffness and damping according to the manipulation loading conditions. Also in this case a virtual mass-spring-damper system is employed, but the main difference between admittance control and impedance control is that the former controls motion after a force is measured, while the latter controls force after motion (or deviation from a set point) is measured \cite{keemink2018}. In \cite{brygo2014a}, the COMAN robot performs free-space motions with compliant limbs to ensure safe interactions during unforeseen collisions on the whole-arm. During the manipulation task, the controller stiffens the humanoid arm joints according to the the external force sensed at the end-effector, permitting to handle the task loads.

Alternatively, the impedance profile can be sent to the robot through a BMI applied to the operator's arm, using for example non-intrusive position and  EMG measurements, technique also known as \textit{tele-impedance} \cite{ajoudani2012}.

\section{Communication Channel}
\label{sec:CommunicationChannel}

In teleoperation, human and humanoid robot transmit information through a \textit{communication channel} \cite{Bemporad_CDC1998}.
However, the communication channel introduces complexities that impact the stability and performance of the teleoperation, namely, transmission delays, and distortion of the information.
		
\subsection{Transmission delays}
		
The interest for the transmission delays in teleoperation emerged back in the 60s.
The first research to consider the effect of these transmission delays on the performance of the teleoperation was published by Ferrel \cite{Ferrel_THFE1965}. 
He realized that when a delay is inserted most operators adopt an effective \textit{move-and-wait} strategy.
However, as he did not consider force reflection there was no problem of instability.
Only when force reflection was considered, instability became apparent.
			
As a consequence, most of the early research on teleoperation targeting delays was based on
supervisory and software-based teleoperation, without	\textit{closing the loop} in force.
Some examples are the employment of high-level commands and stored subroutines,
or a predictive display that gives the operator a
prediction of the system response~\cite{Bemporad_CDC1998}.
			
Although old, these techniques are still in use when dealing with extremely complex robots,
as humanoid robots are.
As an example, the robots that participated during the DRC
had to deal	with a communication channel represented by two links:
\begin{inparaenum}[(a)]
\item a bi-directional low bandwidth (9600 bps) continuous communication path, or
\item a unidirectional high bandwidth (300 Mbps) communication path with intermittent connection.
\end{inparaenum}
The purpose was to boost the autonomy of the robots.
As a result, most of the teams employed high-level commands, visual aids, and no bilateral
teleoperation \cite{feng2015optimization}.
The effect of the \textit{``move-and-wait''}	strategy was apparent, and greatly influenced the operation speed.
			
A real sense of telepresence can only be achieved by	kinesthetically coupling the operator to the remote environment by
introducing force reflection \cite{hokayem2006bilateral}.
However, time delay can represent a destabilizing factor unless the motion bandwidth is severely reduced
or more advanced solutions are considered.
Shared-control solutions can also be adopted to handle communication
delays \cite{liu2013}.
Even round-trip delays around 100~ms (typical of the Internet) in fact, can induce the instability of the teleoperation system either indirectly, due to human operator's overcompensations of the delayed perceived errors \cite{ferrell1965}, or directly in the control law in the case of bilateral systems.

Advanced techniques to deal with delays appeared in the mid 80s.
Particularly, using \emph{network theory} and the concept of passivity allowed to achieve stable
time-delayed teleoperation assuming constant-time delays \cite{Anderson_IJRR1992}~\cite{Niemeyer_JOE1991}.
The \emph{constant-time delay} assumption is limited and valid for simple communication channels.
Packet switched networks, as it is the case of Internet, introduce additional difficulties:
\begin{inparaenum}[(a)]
\item random varying delays that can reach very high values,
\item a discrete-time exchange of information,
\item the effect of quantization, and
\item loss of packets \cite{hokayem2006bilateral}.
\end{inparaenum}
Varying-time delays appear due to several factors like traffic congestion, bandwidth, and information loss.
The latter mainly caused by transmission time-outs, transmission errors, and a limited buffer size.
To deal with the varying-time delay,
one can
estimate an upper bound of the delay
through networks statistics, then use it to dissipate energy \cite{Lozano_Mechatronics2002},
to emulate a virtual constant-time delay \cite{Kosuge_ICRA1996}, or to use it to extend the horizon of a model predictive control \cite{Bemporad_CDC1998}.

\subsection{Distortion}
			
Distortion is another effect introduced by the communication channel.
In case of packet switched networks, a delay of discrete-time velocity information results in a distortion of velocity and position drift, hence, degrading the performance.
Solutions to this problem can include the exchange of position information together with the velocity, or time-stamping the information \cite{Chopra_ACC2003}.
Another source of distortion is due to the policies introduced when there is information loss:
\begin{inparaenum}[(a)]
\item to treat the lost packet as a null packet,
\item to use the last valid packet, or
\item to use interpolation \cite{hokayem2006bilateral}.
\end{inparaenum}
			
\subsection{Network-based Model}
			
The standard model of the communication channel is based on network theory.
By using an analogy between mechanical and electrical systems, we can represent the teleoperation system as
a network of interconnected $n$-ports. 
An $n$-port is characterized by the relationship between \emph{effort} $f$ (voltage or force) and \emph{flow} $v$ (current or velocity).

Assuming that we have flows and efforts on both sides of a 2-port network representing a Linear Time-Invariant
(LTI) system, the relationship between them can be written (in the frequency domain) by a hybrid matrix.
The idea is to use control to modify the characteristics of this matrix in order to overcome the difficulties imposed by the communication channel \cite{hokayem2006bilateral}.
In the topic of bilateral teleoperation represented by a nonlinear system, which is the case for humanoid robots, the frequency-domain approach is no longer valid.
However, Anderson \cite{Anderson_IJRR1992} demonstrated that by using a Hilbert network, with efforts and flows
belonging to a Hilbert space, equivalent analyses could be performed.
			
\subsection{Passivity}
			
The passivity formalism provides a simple and robust tool to analyze the stability of a nonlinear system \cite{Niemeyer_JOE1991}.
A passive system may dissipate energy ($E(t)$) but it cannot increase its total energy \cite{Anderson_TAC1989}.
For an $n$-port, we can take forces as inputs and velocities as outputs, and define the \textit{power}
(not necessarily a physical one) entering to the system as the scalar product between input and output
($P(t) = f^T(t) v(t)$).
Then, for an $n$-port to be passive, the power is either stored or dissipated:
\begin{equation}
\int_0^t f^T(\tau) v(\tau) d\tau = E(t) - E(0) + \int_0^t P_\text{diss} d\tau
\geq -E(0),
\end{equation}
where $P_\text{diss}$ is a non-negative power dissipation function. If no power is dissipated, the $n$-port is lossless \cite{Niemeyer_JOE1991}.

An important tool that can be used to analyze the passivity
is \emph{scattering theory}, which relates the effort and flow of a network through a  \emph{scattering operator}~\cite{Anderson_TAC1989, hokayem2006bilateral}.
In this respect, a passivity-based system imitates a physical system that obeys energy conservation
\cite{Niemeyer_JOE1991}.
An important property of passive systems is that the stored energy and power dissipation of a combined system is equal to the sum of individual stored energies for each system \cite{Niemeyer_JOE1991}.
Therefore, if we assume that the operator and the environment behave passively, the entire system can behave passively if each $n$-port is passive.

\subsection{Stability of Time-delayed Humanoid Robot Teleoperation}
Passivity and stability are related as a result of considering the expression of the stored energy as a Lyapunov function \cite{Niemeyer_JOE1991}.
If the $n$-port corresponding to the communication channel is simply determined by a delay, then it can be demonstrated that a pure delay introduces power~\cite{Anderson_TAC1989}.
However, it is possible to modify the behavior of the communication channel by using control.
A naive solution is to add damping, but there is no stability guarantee, and it affects the performance \cite{Niemeyer_JOE1991}.
Alternatively, Niemeyer et al \cite{Niemeyer_JOE1991} used the power and energy to define the concept of \emph{wave variables}, where input and output of wave variables are a combination of efforts and flows.
Then, a system is passive if the energy of the output waves is less than the energy of input waves.
In a recent work by~\cite{risiglione2021passivity} on bilateral teleoperation of legged robot mobile manipulators, to ensure the passivity and stability they used \textit{energy tanks} as energy observers and included that as a constraint to the robot whole-body control layer.
The problem is that achieving passivity does not guarantee good performance \cite{Chopra_ACC2003}, and stability and transparency are conflicting objectives \cite{hokayem2006bilateral}.

More specifically to the humanoid robot's unilateral and bilateral teleoperation, the delay is relevant to the robot's balance and velocity. These challenges have not yet been addressed in the literature. However, in many applications the delay is low, hence does not affect highly the robot balance through teleoperation. In the case of large delays, a higher autonomy level for teleoperation is required. In the case of mid-range delay, a commonly used strategy is \emph{move-and-wait} to overcome the delay problem, however, this reduces the robot velocity and efficiency in task performance. In the case of bilateral teleoperation and manipulation tasks, the use of this approach is highly limited.
With the rise of machine-learning techniques, an envisioning approach to overcome the delay is to use predictive approaches, both from the human and the robot side.
On one hand, the human motion should be predicted and mapped to the robot, and on the other hand, the robot motion and interaction forces with the environment, as well as the stream of the visual feedback, should be predicted and sent to the human operator. 
Finally, to speculate the humanoid robot's balance through teleoperation with a lower level of autonomy, we may consider the human operator, and retargeting and control strategy, all together, as a complex control system to balance the humanoid robot. In this case, disturbances on the robot can be controlled through the whole teleoperation pipeline and human commands.
By speculating on the simplified robot model, i.e., LIPM , studies shows that if the time delay of the system is higher than a critical time delay, the robot destabilizes \cite{milton2009time}. The critical time delay ${\tau}_{c}$ is relevant to the robot CoM height, i.e., ${\tau}_{c} = \beta  \sqrt{\frac{z_0}{g}}$ from Eq.~\eqref{eq:lipm}. The lower the CoM height, the lower becomes ${\tau}_{c}$.
This is related to the fact that a lower CoM height implies faster dynamics of the LIPM, hence a lower time delay is tolerated to keep the humanoid robot's balance. Here $\beta$ can be related to different parameters such as the human performance, the retargeting and control strategy, and the robot actuators.

\section{Design \& Evaluation of Humanoid Teleoperation System}
\label{sec:EvaluationMetrics}

In order to deploy the teleoperation system for \textit{real users}, such systems should meet the users' needs and be evaluated prior to deployment. These needs and considerations should be anticipated in the design and development phase.
A human-centered design, consisting in an iterative process aiming at properly allocating the tasks (functions) between the user and the teleoperation system, is required.
Designers need metrics to evaluate the teleoperation systems, to share the knowledge, and compare their findings. These metrics are highly task-dependant, i.e., various tasks impose different functional requirements that the system should meet \cite{Steinfeld2006Common, Goodrich2007Survey}.
These metrics not only determine inherent problems and limitations of the humanoid robot teleoperation system, but also provide guidelines for the design and development, reducing the cost and the time to design such a system.
The rest of this section adopts some metrics proposed in relevant fields that help design and evaluate humanoid robot teleoperation systems.

\subsection{Evaluation Metrics}

\subsubsection{Usability Assessment}
According to ISO 9241 \cite{ISO9241}, \textit{usability} is defined as \textit{``the extent to which a system, product or service can be used by specified users to achieve specified goals with effectiveness, efficiency and satisfaction in a specified context of use''.}
In line with this definition and the teleoperation context, effectiveness is the degree of accuracy and completeness in which the user reaches the teleoperation goal, whereas efficiency is related to the resources being used (e.g., time, cost, human effort, etc.) with respect to the achieved outcome \cite{ISO9241}. 
Effectiveness and efficiency are considered as the objective usability measures, which depend on the teleoperation task.
On the other hand, satisfaction determines the degree in which the users' needs and expectations are met as a result of the use of the teleoperation system \cite{ISO9241}. It is determined according to the user's emotional, physical, and cognitive responses.
The user's perceived usability,
which is a subjective measure,
has a direct relation with the user's satisfaction; the more the perceived usability, the higher the satisfaction \cite{Flavian2006Role}.
Moreover, usability assessment is relevant to the individual user in terms of the frequency of use and familiarity with the system.
Some criteria to measure the effectiveness of a task execution are the task completeness, objective achievement, and task errors \cite{Bevan2016New}.
The efficiency of a task execution is established according to the task time, cost-effectiveness, energy consumption, productive time ratio, and unnecessary actions (just a few to mention) \cite{Bevan2016New}.
For the subjective measure of the usability there are some standard empirical and informal techniques to examine the user's perceived usability including concurrent think aloud, retrospective think aloud, concurrent probing, and retrospective probing.
Among these, the most famous one is the System Usability Scale (SUS) \cite{Brooke1996SUS} retrospective probing technique which assesses two factors including  usability as well as learnability for a variety of tasks. Nevertheless, a subjective measure of the usability is in relation with the user perception; therefore, it is affected by the situational awareness, which will be discussed in the next section.
An attempt to provide a holistic taxonomy of usability evaluation in robot teleoperation scenarios is done in \cite{Adamides2014Usability}.

\subsubsection{Situational Awareness (SA)}

SA correlates the user capabilities, training, experiences, preconceptions, and objectives with the ongoing task workload together~\cite{endsley1988situation}.
Endsley identified \textit{situation awareness} with three levels as \textit{``the perception of elements in the environment within a volume of time and space, the comprehension of their meaning, and the projection of their status in the near future"}~\cite{endsley1988situation}.
Accordingly,
SA 
tries to understand the process which leads to decision making, considering a variety of elements.

High SA promotes the probability of a good performance, and the poor performance of the task is normally the result of SA loss. SA is lost when the user's knowledge is incomplete or inaccurate, the user attention is narrowed to some elements, or when the mental model of the user diverges from the reality~\cite{endsley1988situation, Nguyen2019Review}.
Loss of SA is also correlated to the workload; when the workload is high the user's attention is drawn from the main task and therefore the user does not give importance to the rest of the elements.

The main works in the literature measuring SA are based on the probe technique, physiological measurements, implicit inference, process indices, and observer ratings~\cite{endsley1988situation, Nguyen2019Review}. Subjective ratings of the SA, where the users fill a form,
are limiting because of the inaccurate evaluation of the users about their SA and the biased evaluation depending on the task outcome~\cite{endsley1988situation}.
To assess SA, EEG and eye-tracking physiological signals are employed as well in the literature. 
SA can be inferred indirectly and implicitly through other related measurable metrics, like the task performance~\cite{Nguyen2019Review}.
Finally, probe techniques can be both retrospective, after task execution, or concurrent, by collecting data with a questionnaire while executing the task.
Among the concurrent ones, the Situation Awareness Global Assessment Technique (SAGAT) proposed by Endsley in~\cite{endsley1988situation} is a freeze-probe approach that measures SA objectively.
This method interrupts the experiment at some random points and halts the user interface,
then a number of questions are asked from the users to evaluate their knowledge about the current and future situation including the user's perception, comprehension, and actions. Later, the responses are compared with the real values collected from the scenario or from experts' responses.
While SAGAT is very reliable, it is intrusive to the natural flow of the task execution.

\subsubsection{Workload}

Workload relates the resources demanded by a task to the available resources supplied by the human operator.
There is no consensus about the definition of the workload \cite{Miller2001Workload};
however, \cite{Hart1988Development} identifies the \textit{workload} as \textit{``the amount of work that is loaded on an individual, the time pressure in which a task is performed, the level of effort exerted, the success in meeting the task requirements, physiological and psychological"}. 
Workload is related to the human operator's (subjective) experience in response to the task objectives.
Especially in time critical decision making tasks, a high workload can lead to user errors or to a delay in the decision making.

Workload presents high variability depending on the human operator and the teleoperation task, hence being difficult to measure.
In fact, the workload is multi-dimensional,
and various subject- and task-related factors (such as mental, physical, information, perceptual, and communication loads) should be considered~\cite{ Miller2001Workload, Hart1988Development}.
Different approaches have been proposed in the literature for measuring the workload on the human operator, namely questionnaires and experts' reports (subjective), physiological techniques, and performance-based approaches~\cite{Miller2001Workload}.
Specifically, for a reliable mental workload assessment a mixture of them is suggested in~\cite{Miller2001Workload}.
Among the subjective measurement methods, the NASA Task Load Index (NASA-TLX) and the Subjective Workload Assessment Technique (SWAT) are the most famous multi dimensional subjective rating methods used in the literature~\cite{Hart1988Development, Reid1988Subjective}. 
Continuous physiological measurements such as cardiac activity, eye activity, respiratory activity, speech measures, and brain activity can provide a reliable assessment of the task physical or mental load as well  \cite{Miller2001Workload}
To measure the physical workload, oxygen consumption estimation can directly provide an index, while heart rate can be used as an indirect method.
Moreover, since the workload and task performance have a causal relationship, the task execution performance can play as another indicator of the workload, when time-shared or difficulty of tasks are exploited. An extensive review of the methods to measure the workload can be found in~\cite{Miller2001Workload}.

\subsubsection{Engagement, Immersion, Involvement, and Presence}
Some metrics concerning robot teleoperation, specifically related to the subjective experience of the user are engagement, immersion, flow, involvement, and presence.
In different fields, engagement is defined and measured distinctly. In the context of teleoperation, the definition and measurement approaches of engagement are especially relevant in gaming and virtual reality applications.
According to~\cite{bouvier2014defining}, engagement is \textit{``the willingness to have emotions, affect, and thoughts directed towards and aroused by the mediated activity in order to achieve a specific objective''.}
It relies on the user’s activity and expectations. The user is engaged when her/his perceptual, intellectual, and interactional expectations are met.
In this context,  \textit{presence} is characterized as \textit{``the subjective experience of being in one place or environment, even when one is physically situated in another''}  \cite{witmer1998measuring}. Presence is a multifaceted concept that is related to \textit{involvement}, a psychological state  depending on attention to remote environment stimuli, and \textit{immersion}, a psychological state of perceiving oneself as a part of the remote environment stimulus flow \cite{witmer1998measuring}.
Here, stimulus flow is a dynamic stream of sensory information.
There are several factors contributing toward the sense of presence including control, sensory information, distraction, and realism factors.
More information about \textit{presence} can be found in~\cite{witmer1998measuring}.

Even if presence is a subjective sensation, there are both subjective and objective methods to measure it. Subjective measures are based on self-report questionnaires like the ITC-sense of presence inventory \cite{lessiter2001cross} or the Witmer \& Singer Presence Questionnaire \cite{witmer1998measuring}. For objective measures, both behavioral (e.g., startle response) and physiological measures (e.g., heart-rate changes, skin conductance/temperature) are suggested in the literature \cite{riva20037}.
Similar methods have also been exploited to measure the user engagement \cite{rani2005operator}.

\subsection{Interface Design and Human-centered Autonomy}

The interface design impacts the user's decisions while teleoperating the humanoid robot, being the medium of communication between human and robot. Good design choices leverage the user experience and result in a successful teleoperation system.
The diagnostic information provided by different measurement techniques can eventually enhance the design and the user support.
Following the explanation of different metrics provided before, we can identify four types of measurement methods, namely, questionnaire-based subjective ratings, performance-based objective ratings, physiological measures, and behavioral responses.
The subjective rating evaluation is quick, and can be done either online (while performing the task) or retrospectively.
Nevertheless, studies show that retrospective evaluation, while not being intrusive, may be subject to information loss because of the user's bias and poor subject ability to recall.
Physiological measurements are partly related to the concept of \textit{activation} or \textit{arousal} in the field of psychophysiology, which relates the changes of mental effort and its effect on the physiological activity of the human operator~\cite{Roscoe1992Assessing}.
Finally, all the metrics provided are interrelated; for example,~\cite{Riley2004Situation} has studied the effect of SA and task difficulty level on the human performance and the user's sense of telepresence.

The described metrics can be used to identify the robot behaviors that should be automated.
When designing a teleoperation system, one of the critical parameters that impacts the success of the system and the user experience is the autonomy level.
It affects the SA, the user workload, the system usability and the user's sense of presence~\cite{Deng2019Influence}.
While high degree of automation degrades the SA, low autonomy in complex tasks intensifies the workload.
High level of autonomy promotes the user's performance and diminishes the workload in the normal situations~\cite{Kaber2000Design}, while during system failures, a low level of autonomy increases the SA and enhances the human manual performance.
Therefore an intermediate level of autonomy in which the user is in the control loop, i.e., human-centered autonomy, enhances the SA, reduces the workload, and improves the performance against failures~\cite{Kaber2000Design}.

\subsection{ Humanoid Robot Teleoperation Design, Evaluation Challenges \& Future Directions}

The design of humanoid robots and teleoperation systems are inherently iterative, where the system goes under different evaluations. However, to improve the design, not many studies in the literature have been published to evaluate such systems systematically.
Table \ref{tab:metrics} provides an overview of the works where different metrics are used to enhance and evaluate the teleoperation system. This table demonstrates that fewer behavioral and physiological measurements are employed to evaluate the developed system. We speculate that this is due to the lack of competencies in those fields in the robotics community.
More recently, ANA Avatar XPRIZE \cite{AnaAvatarXprize} took the lead in the evaluation of humanoid robot teleoperation systems by assessing both task performance and subjective measures in locomotion, manipulation, and social interaction scenarios. 
These evaluations can help the developers to choose proper teleoperation devices and adjust the autonomy level, more specifically retargeting and control approaches for humanoid teleoperation according to the architecture presented in Sec.~\ref{fig:retargeting_controller_architecture}.

\begin{table}
\setlength\tabcolsep{5pt}
\centering
\arrayrulecolor{black}
 \caption{Examples of evaluation metrics for the design and evaluation of robot teleoperation setups.}
\begin{tabular}{|c!{\color{black}\vrule}c|c|c|c|c|c|c|c|c|} 
\hline
\multirow{2}{*}{{ref}}   & \multicolumn{4}{c|}{evaluation metrics}                  & ~                     & \multicolumn{4}{c|}{measurements}   \\ 
\cline{2-5}\cline{7-10}
                                        & {\rotatebox[]{-90}{\makecell{usability}}} & {\rotatebox[]{-90}{\makecell{situational \\ awareness}}} & {\rotatebox[]{-90}{\makecell{workload}}} & {\rotatebox[]{-90}{\makecell{engagement, ~ \\ presence}}} & {\rotatebox[]{-90}{\makecell{ autonomy attentive~ \\ system}}} & {\rotatebox[]{-90}{\makecell{subjective \\ measures}}} & {\rotatebox[]{-90}{\makecell{objective \& task \\performance}}} & {\rotatebox[]{-90}{\makecell{physiological}}} & {\rotatebox[]{-90}{\makecell{behavioural}}} \\ 
\hline\hline

\rowcolor[rgb]{0.906,0.902,0.902}
\cite{Kaber2000Design}        & ~         & \checkmark     & \checkmark        & ~          &  \checkmark         &  \checkmark         &  \checkmark         & ~         & ~          \\

\cite{Lu2019Workload}         & \checkmark            & ~     & \checkmark        & ~          & ~         & \checkmark         & \checkmark         & \checkmark          & ~         \\

\rowcolor[rgb]{0.906,0.902,0.902}
\cite{Riley2004Situation}                                       & ~         & \checkmark     & ~         & \checkmark          &  ~      & \checkmark         & \checkmark         & ~          &  ~         \\

\cite{rani2005operator}       & ~         & ~         & ~        & \checkmark          & \checkmark         & \checkmark         & ~         & \checkmark          & ~         \\

\rowcolor[rgb]{0.906,0.902,0.902}
\cite{Radmard2015Interface}   & \checkmark            & ~     & \checkmark        & ~          & ~         & \checkmark         & \checkmark         & ~          & ~         \\

\cite{Jankowski2015Usability}                                   & \checkmark            & ~     & ~        & \checkmark          & ~         & \checkmark         & \checkmark         & ~         & ~          \\

\rowcolor[rgb]{0.906,0.902,0.902}
\cite{Singh2018Physiologically}                                 & ~         & ~         &   \checkmark       & ~          & \checkmark         & \checkmark         & ~         & \checkmark          & \checkmark       \\

\cite{Barros2011Enhancing}                                      & ~         & \checkmark & ~    & ~          & ~         & \checkmark         & \checkmark         & ~          & ~         \\

\rowcolor[rgb]{0.906,0.902,0.902} 
\cite{Villani2017Natural}     & \checkmark            & ~     & ~        & ~          & ~         & ~         & \checkmark         & ~          & ~         \\

\hline
\end{tabular}
\arrayrulecolor{black}

     \label{tab:metrics}
\end{table}

While developing a humanoid robot teleoperation system, different hardware and software components are interleaved in order to perform a task. In the case of poor performances, it is important to consider the interconnection of different components and try to find the sources of the problems.
Another challenge related to teleoperation system development might be latency, which can be related to perception, communication, control, or actuation.
An experimental demonstration provided by~\cite{Lu2019Workload} showed a detrimental impact of the time delay
to the human operator's increased workload as well as the performance. Moreover, the discrepancy of the time delay among different components, e.g., arms, torso, locomotion, hand, and social cues (such as facial expression), can further degrade the teleoperation performance and the user's experience, highlighting the importance of synchronized and coordinated motion, and social cues for the teleoperated humanoid robot.
Unique to humanoid robot teleoperation, having an anthropomorphic humanoid robot body, and a proper selection of teleoperation interfaces, algorithms, and low latency can lead to strong telepresence, or so-called tele-embodiment~\cite{paulos2001social}.

A final challenge in humanoid teleoperation is to build a system that can be easily mastered by non-expert users. State of the art does not often consider any usability assessment or carry out any user study, strongly relying on the expertise and training of a single human operator. From this perspective, we hope that this section helps the reader in identifying the key points and right evaluation metrics for deploying a teleoperation system that meets the user's needs.

\section{Applications and Perspectives}
\label{sec:Applications}

Future applications of teleoperated humanoid robots are very promising but further progress is required to employ these robots in real world scenarios. 
For successful teleoperation, the theoretical foundations are provided in the previous sections.
Here, we overview major applications of the humanoid robot teleoperation, and discuss their challenges and opportunities.

\subsection{Telexistence and Telepresence}
The COVID-19 pandemic has either canceled many conferences and meetings around the world or led them to be transformed into virtual events.
Moreover, many people cannot see their beloved ones frequently. The current substitution for such cases currently is to use communication mediums, allowing the people to see and hear each other. However using these means, many social cues -equally important in social interaction- are not yet conveyed. Instead, humanoid robot teleoperation with anthropomorphic shape and motion, similar to a human, would allow for a better experience. Moreover, it would allow people to interact physically with each other.
Currently, the ANA Avatar XPRIZE competition \cite{AnaAvatarXprize} is aiming at this application. 
The main challenges in this context are to allow the operator and the humans who interact with the robot to engage in a scenario, feel the presence of the operator, and enable manipulation in the remote environment.

\subsection{Teleoperation in Hazardous Environments}
One of the main applications urging robot teleoperation is in disaster-response scenarios where the environment is potentially hazardous to humans.
This need arose, for example, in the Chernobyl and Fukushima Daiichi Nuclear Power Plants crises \cite{nagatani2013emergency}. As a response to these disasters, robots were needed to perform surveillance, search and rescue, and manipulation tasks. 
However, to do that, a robot should reach the point of interest passing possibly through a dynamic and unstructured environment to perform a given task.
Bipedal locomotion overcomes other locomotion means in terms of mobility, agility, and a wide range of motion of the robot; therefore, reliable humanoid robot teleoperation can assist the human in such cases.
In these scenarios, the mission requirements allow to identify necessary robot features such as hardware reliability, communication protocols, additional hardware and sensors, autonomy level, and so on.
Normally, in natural disasters the time is very critical; therefore, another important aspect
to consider
is the response time and the technology readiness level. For example, in the Fukushima 
crisis where the human could not enter because of the radiation levels, it took more than three months to deliver a robot that could perform the first mission \cite{nagatani2013emergency}. However, in cases such as Urban Search and Rescue Tasks, where humans' lives are jeopardized, 
the delay for intervention is not admissible~\cite{casper2003human}.

\subsection{Teleoperation in Manufacturing \& Research Environments}

Humanoid robots, despite being attractive because of their anthropomorphism and potential to operate in environments designed for humans, are still expensive and not affordable for many industries and research institutes.
So far, this inconvenience has strongly limited their study to a restricted robotics community.
Moreover, very often either the robot hardware or the underlying control software (or both) are not designed to comply with control failures, loss of balance or wrong interactions with the environment, which can lead to costly breakage. 
In an example of teleoperation in a manufacturing environment by Kheddar et. al. by \cite{kheddar2019humanoid}, the humanoid robots TORO and HRP-4 have been deployed in an aircraft manufacturing environment for assembly operations.
Another example in construction sites can be found in~\cite{yokoi2003tele}.

\subsection{Telenursing}
Front-line healthcare workers are exposed to infectious diseases and are at higher risk of infection compared with the general community. This was evident during disease outbreaks, as experienced during the recent COVID-19 pandemic.
Nurses physically interact with their patients and the environment for a wide variety of tasks such as manipulating objects, taking measurements, and interacting with them.
In this respect, teleoperated robots can facilitate the situation and improve the nurses' safety by performing some of their tasks, exemplified in~\cite{li2017development}.
However, robots should not only perform the nurses' tasks but also avoid patients' discomfort or avoid limiting other health worker activities.
In this respect, the teleoperation of humanoid robots can potentially overcome these challenges when being deployed in clinical environments with the supervision of professional operators and other healthcare personnel.

\subsection{Space Applications}
Space robotics has many applications, including satellite on-orbit servicing, maintenance of the ISS, performing experiments there, and interplanetary exploration and construction~\cite{workshop2019, flores2014review}.
Some of these tasks cannot be performed by humans due to cost, safety, and the increased complexity of the required system. 
On the other hand, the reliability and robustness of autonomous robots are not yet sufficient to perform such tasks autonomously. Therefore, teleoperation of space robots with different degrees of autonomy is required.
Space applications are more challenging due to communication latency and bandwidth, possible unknown kinematics and dynamics properties of the target objects of manipulation, and human factors~\cite{flores2014review}.
In the case of bilateral teleoperation, the round trip communication delay matters; however, time-domain passivity control approaches can compensate to some extent for earth-orbiting robots with the cost of degrading the efficiency \cite{ryu2010passive}.
However, for application with higher distances, it is a compromise between autonomy, risk, and efficiency.
To overcome that, \cite{Lii2017Toward} proposed a supervised autonomy framework with a natural interface to astronauts in the ISS for teleoperating the SUPVIS Justin robot on Earth, simulating an interplanetary solar panel service and maintenance task.

\subsection{Service Robotics Application}
Another area of use of humanoid robot teleoperation is in domestic environments such as houses,
supermarkets, schools, and hotels with a diverse range of goals such as giving care to elderly people or housekeeping~\cite{broekens2009assistive}, restocking the market shelves,
teleducation, guiding visitors in hotels, or teletourism. 
These environments are intrinsically unstructured and built for humans' use; therefore, a humanoid robot is likely to be deployed for such applications.
These applications will become more evident when the operator of the humanoid robot cannot be present in the target environment. In other words, workforces who teleoperate the humanoid robot can be at any place.
The main requirements for such applications to be acceptable are the safety of humanoid robots with the people whom they are interacting with, and the ability to manipulate and modify remote locations.

\vspace{-0.2cm}


\bibliography{references}


\begin{IEEEbiography}[{\includegraphics[width=1in,height=1.25in,clip,keepaspectratio]{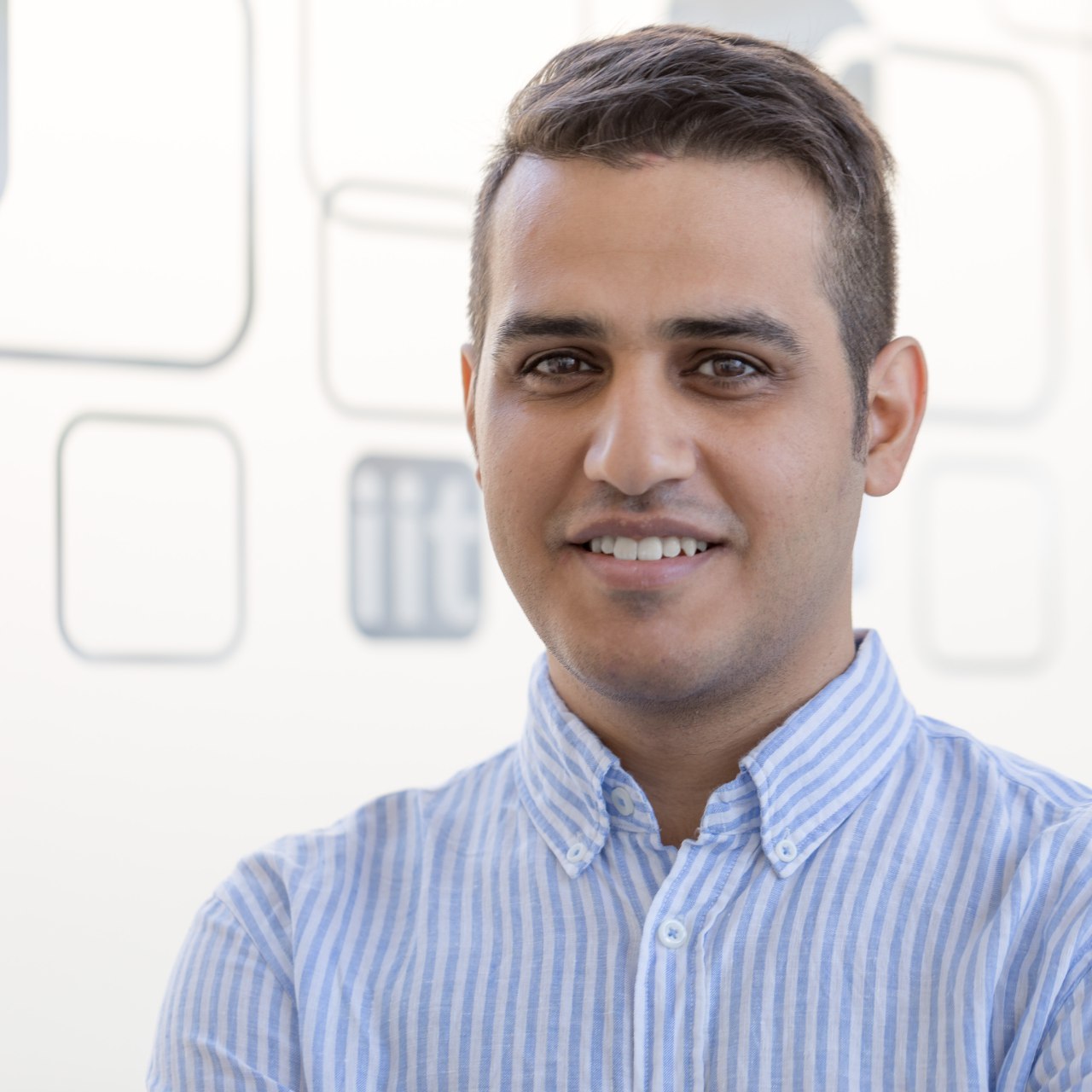}}]{Kourosh Darvish}
Kourosh Darvish is a postdoctoral researcher at the Computer Science and Robotics Institute of the University of Toronto (UofT) and a member of the Vector Institute.
Before joining UofT in 2022, he was a postdoctoral researcher at the Italian Institute of Technology (IIT).
In 2019, he completed his PhD in Bioengineering and Robotics from the University of Genoa, Italy.
His research focuses on shared autonomy, teleoperation, human-robot collaboration, and humanoid robotics.

\end{IEEEbiography}

\begin{IEEEbiography}[{\includegraphics[width=1in,height=1.25in,clip,keepaspectratio]{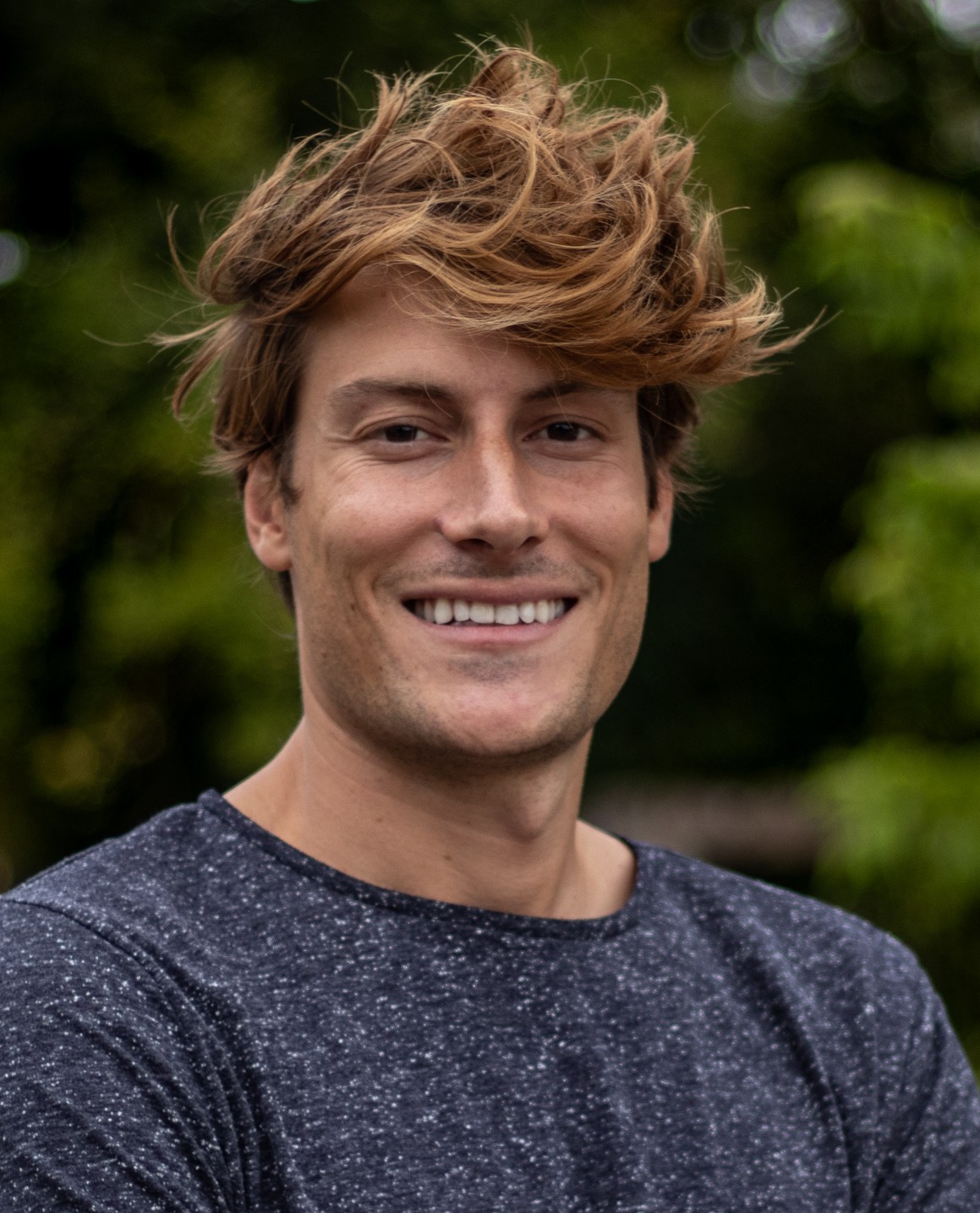}}]{Luigi Penco}
Luigi Penco is a Senior Research Associate at the Florida Institute for Human and Machine Cognition (IHMC).
He earned his bachelor’s degree in Electronics Engineering from Roma Tre University in 2015 and a master’s in Artificial Intelligence and Robotics from La Sapienza University of Rome in 2018. In 2022 he received a PhD in robotics from Université de Lorraine, while conducting his doctoral studies at Inria Nancy Grand-Est.
His research focuses on humanoid robotics, with a particular interest in teleoperation and machine learning techniques used to improve the control and skills of robots.



\end{IEEEbiography}

\begin{IEEEbiography}[{\includegraphics[width=1in,height=1.25in,clip,keepaspectratio]{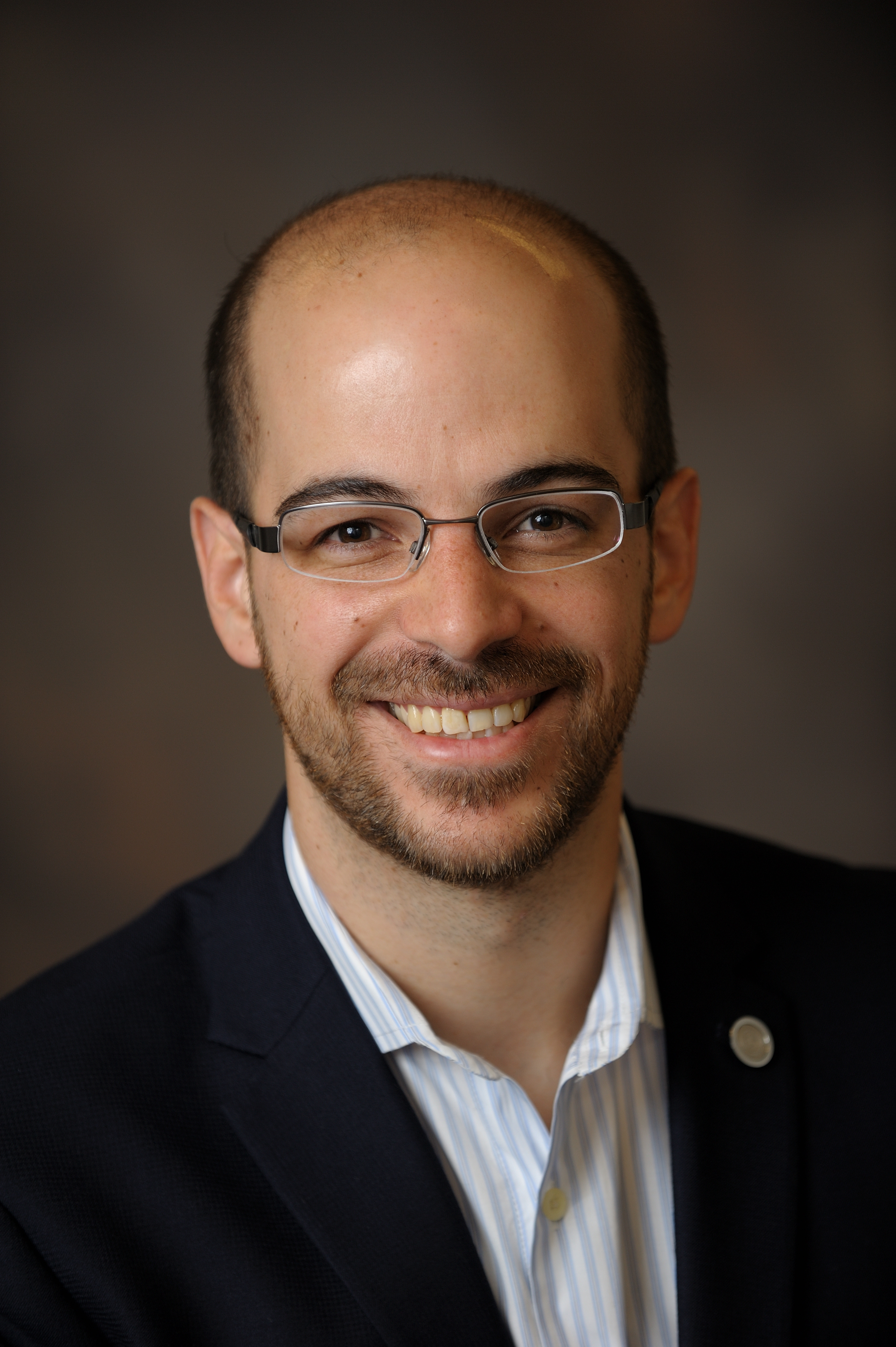}}]{Joao Ramos}
 Joao Ramos is an Assistant Professor at the University of Illinois at Urbana-Champaign (UIUC) and the director of the RoboDesign Lab. He previously worked as a Postdoctoral Associate at the Biomimetic Robotics Laboratory at MIT before joining UIUC in 2019. He received a PhD from the Department of Mechanical Engineering at MIT in 2018. He is the recipient of the 2021 NSF CARRER Award. His research focuses on the design and control of dynamic humanoid robots, human-machine interfaces for whole-body teleoperation, legged locomotion dynamics, and actuation systems.
\end{IEEEbiography}

\begin{IEEEbiography}[{\includegraphics[width=1in,height=1.25in,clip,keepaspectratio]{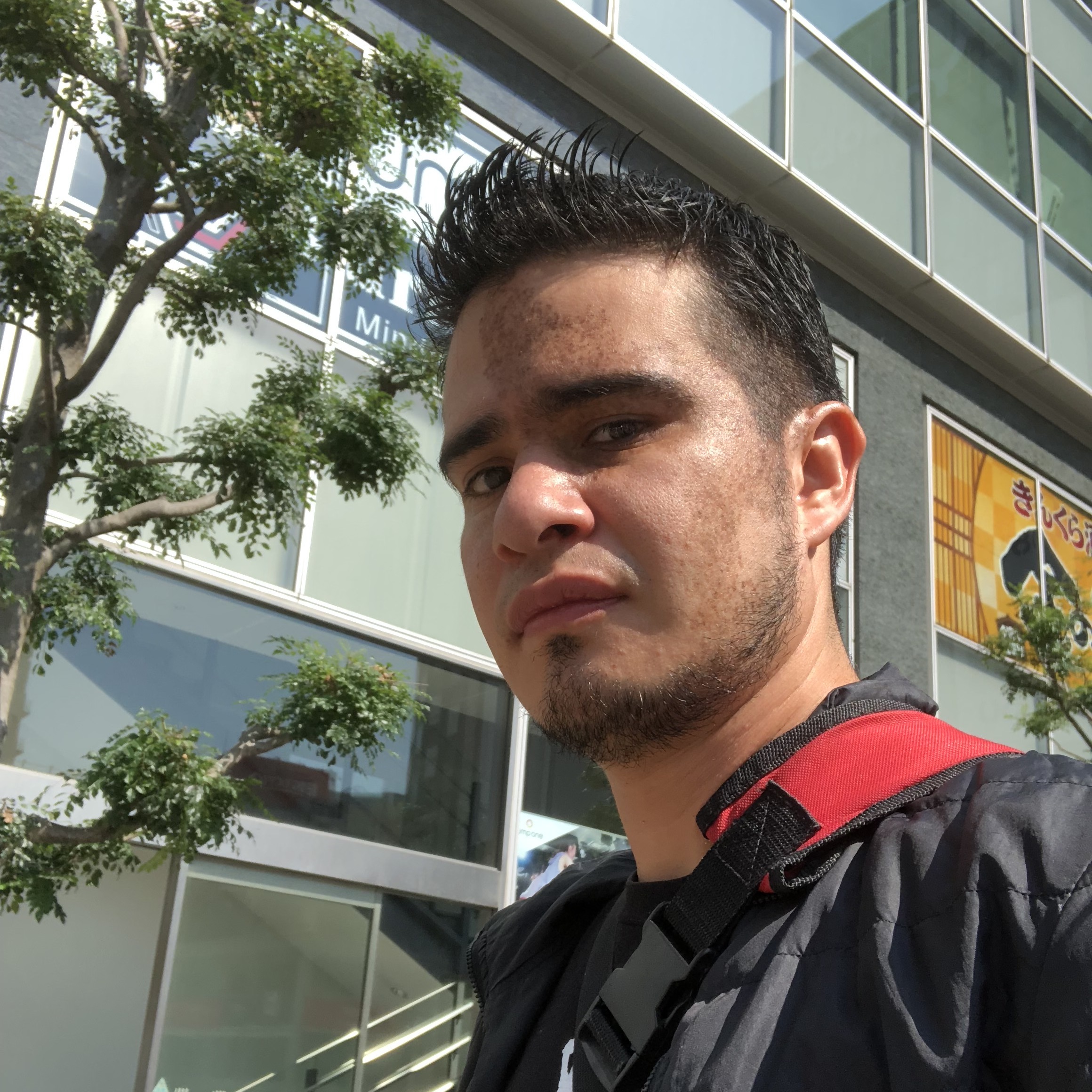}}]{Rafael Cisneros}
Rafael Cisneros (Member, IEEE) received the B.Eng. degree in Electronics and Computers from the University of the Americas—Puebla (UDLA-P), Puebla, Mexico, in 2006, the M.Sc. degree in Automatic Control from the Center of Research and Advanced Studies, National Polytechnic Institute (CINVESTAV-IPN), Mexico City, Mexico, in 2009, and the Ph.D. degree in Intelligent Interaction Technologies from the University of Tsukuba, Tsukuba, Japan, in 2015. Since then, he has been with the National Institute of Advanced Industrial Science and Technology (AIST), Tsukuba, Japan, from 2015 to 2018 as a Postdoc, and since 2018, as a Researcher. He is currently a member of CNRS-AIST JRL (Joint Robotics Laboratory), IRL, AIST. His research interests include torque control, whole-body multi-contact motion control of humanoid robots, multibody collision dynamics, teleoperation, and tactile feedback.
\end{IEEEbiography}

\begin{IEEEbiography}[{\includegraphics[width=1in,height=1.25in,clip,keepaspectratio]{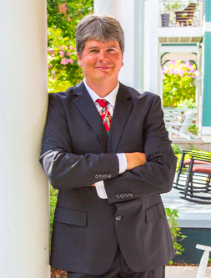}}]{Jerry Pratt}
Jerry Pratt is the CTO of Figure, a startup creating general purpose humanoid robots for commercial applications. Before joining Figure, Jerry was the Co-founder of Boardwalk Robotics, as well as a Principal Investigator at IHMC, where he co-led a research group focused on walking and running robots and exoskeletons. Jerry was the team lead for Team IHMC in the DARPA Robotics Challenge (DRC). In 2015 IHMC won second place in the DRC finals. 
\end{IEEEbiography}

\begin{IEEEbiography}[{\includegraphics[width=1in,height=1.25in,clip,keepaspectratio]{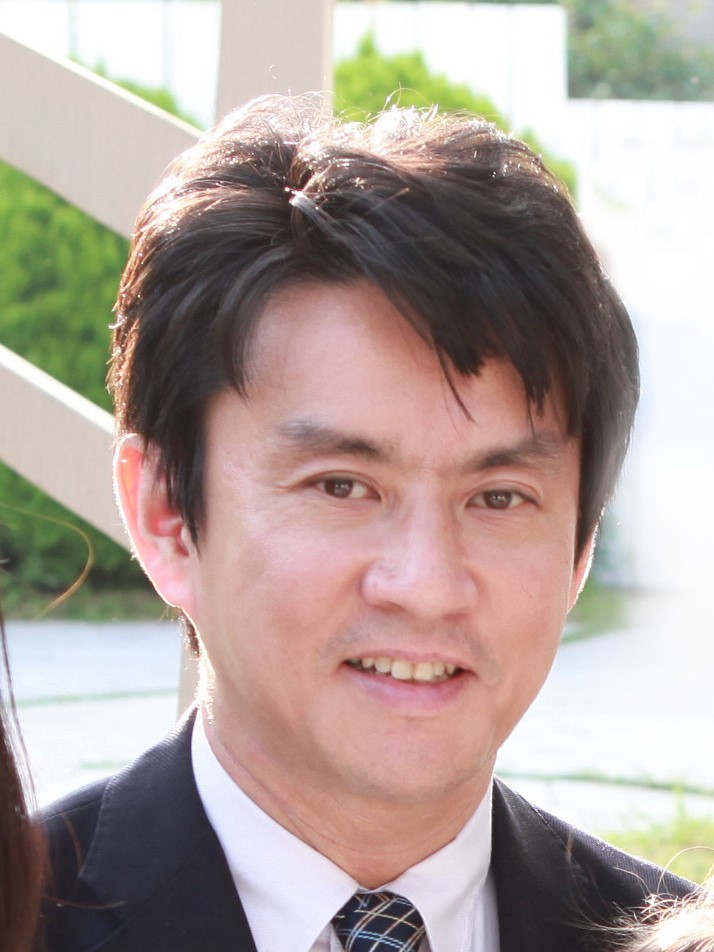}}]{Eiichi Yoshida}
Eiichi Yoshida received M.E. and Ph.D. degrees on Precision Machinery
Engineering from Graduate School of Engineering, the University of
Tokyo in 1996. He then joined former Mechanical Engineering
Laboratory, later in 2001 reorganized as National Institute of Advanced
Industrial Science and Technology (AIST), Tsukuba, Japan. He served as
Co-Director of AIST-CNRS JRL (Joint Robotics Laboratory) at LAAS-
CNRS, Toulouse, France, from 2004 to 2008, and at AIST, Tsukuba,
Japan from 2009 to 2021.
Since 2022, he is Professor of Tokyo University of Science, at Department of Applied Electronics, Faculty of Advanced Engineering. He is IEEE Fellow, and member of RSJ, SICE and JSME and is currently
serving as Senior Editor of IEEE Transactions on Robotics.
His research interests include robot task and motion planning, human modeling, humanoid robotics and advanced logistics
technology.
\end{IEEEbiography}

\begin{IEEEbiography}[{\includegraphics[width=1in,height=1.25in,clip,keepaspectratio]{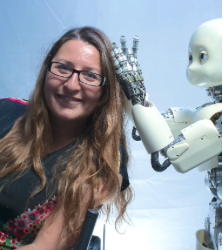}}]{Serena Ivaldi}
Serena Ivaldi is a research scientist at Inria Nancy Grand-Est, France. She obtained her Ph.D. in Humanoid Technologies in 2011 at the Italian Institute of Technology and University of Genoa, Italy, and the French Habilitation to Direct Research (HDR) at the University of Lorraine, France, in 2022. Her research is focused on humanoid robotics and human-robot interaction. 
\end{IEEEbiography}

\begin{IEEEbiography}[{\includegraphics[width=1in,height=1.25in,clip,keepaspectratio]{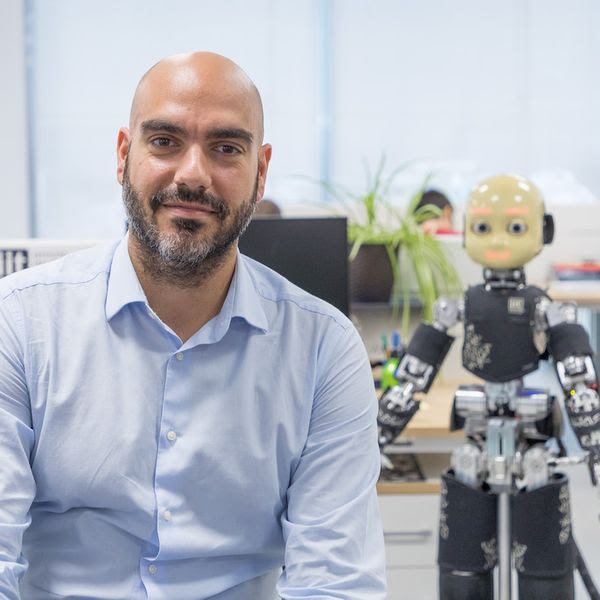}}]{Daniele Pucci}
Daniele received the bachelor and master degrees in Control Engineering with highest honors from ”Sapienza”, University of Rome, in 2007 and 2009, respectively. In 2013, he earned the PhD title with a thesis prepared at INRIA Sophia Antipolis, France, with the supervision of Tarek Hamel, Salvatore Monaco, and Claude Samson. From 2013 to 2017, he has been a postdoc at the Istituto Italiano di Tecnologia (IIT).
From August 2017 to August 2021, he has been the head of the Dynamic Interaction Control lab.
Since September 2021, Daniele is the PI leading the  Artificial and Mechanical Intelligence  research line at IIT, a team composed of about forty members that combines AI and Mechanics to devise the next generation of the iCub humanoid robot. 
\end{IEEEbiography}

\end{document}